\documentclass[11pt]{article}

\usepackage[preprint]{acl}

\usepackage{times}
\usepackage{latexsym}
\usepackage{booktabs} 
\usepackage{multirow} 
\usepackage{graphicx} 
\usepackage{rotating}   
\usepackage[most]{tcolorbox}
\usepackage[table]{xcolor}
\usepackage{graphicx}
\usepackage{subcaption}
\usepackage{enumitem}

\usepackage[utf8]{inputenc}
\usepackage[T1]{fontenc}
\usepackage{amsmath}
\usepackage{amssymb}
\usepackage{algorithm}
\usepackage{algpseudocode}

\usepackage[T1]{fontenc}

\usepackage[utf8]{inputenc}

\usepackage{microtype}

\usepackage{inconsolata}

\usepackage{graphicx}
\usepackage{amsmath}
\usepackage{amssymb}
\usepackage{tabularx}
\usepackage{xcolor}
\definecolor{f1Green}{RGB}{138, 169, 113}

\newcommand{\ourmethod}{{\fontfamily{ppl}\selectfont MedEinst}}

\title{\ourmethod: Benchmarking the Einstellung Effect in Medical LLMs\\ through Counterfactual Differential Diagnosis}

\author{
    Wenting Chen$^{1}$\thanks {Equal contribution.},
    Zhongrui Zhu$^{2}$\footnotemark[1], 
    Guolin Huang$^{3}$, 
    Wenxuan Wang$^{4}$\thanks{Corresponding Author.}\\
    $^{1}$Stanford University, 
    $^{2}$Xi'an Jiaotong University, \\
    $^{3}$Shenzhen University, 
    $^{4}$Renmin University of China
}

\begin{document}
\maketitle
\begin{abstract}
Despite achieving high accuracy on medical benchmarks, LLMs exhibit the Einstellung Effect in clinical diagnosis—relying on statistical shortcuts rather than patient-specific evidence, causing misdiagnosis in atypical cases. Existing benchmarks fail to detect this critical failure mode. We introduce \textbf{\ourmethod}, a counterfactual benchmark with 5,383 paired clinical cases across 49 diseases. Each pair contains a control case and a "trap" case with altered discriminative evidence that flips the diagnosis. We measure susceptibility via Bias Trap Rate—probability of misdiagnosing traps despite correctly diagnosing controls. Extensive Evaluation of 17 LLMs shows frontier models achieve high baseline accuracy but severe bias trap rates. Thus, we propose \textbf{ECR-Agent}, aligning LLM reasoning with Evidence-Based Medicine standard via two components: (1) Dynamic Causal Inference (DCI) performs structured reasoning through dual-pathway perception, dynamic causal graph reasoning across three levels (association, intervention, counterfactual), and evidence audit for final diagnosis; (2) Critic-Driven Graph \& Memory Evolution (CGME) iteratively refines the system by storing validated reasoning paths in an exemplar base and consolidating disease-specific knowledge into evolving illness graphs. Source code is to be released.

\end{abstract}



\section{Introduction}

Large Language Models (LLMs) \citep{achiam2023gpt,llama2} and LLM-based agents \citep{medagents,kim2024mdagents} achieve high performance on medical benchmarks~\citep{medqa}. However, \citet{kim2025limitations} show these models exhibit the \textbf{Einstellung Effect}, relying on statistical shortcuts rather than logical reasoning. This causes models to prioritize common patterns over patient-specific evidence when encountering misleading features, ignoring key discriminative evidence. This effect is particularly problematic in differential diagnosis (DDx), where distinguishing between competing hypotheses depends on subtle symptomatic differences. Mitigating the Einstellung Effect in DDx is essential for deploying trustworthy clinical AI systems.

\begin{figure}[t]
    \centering
    \vspace{-0.2cm} 
    \includegraphics[width=\linewidth]{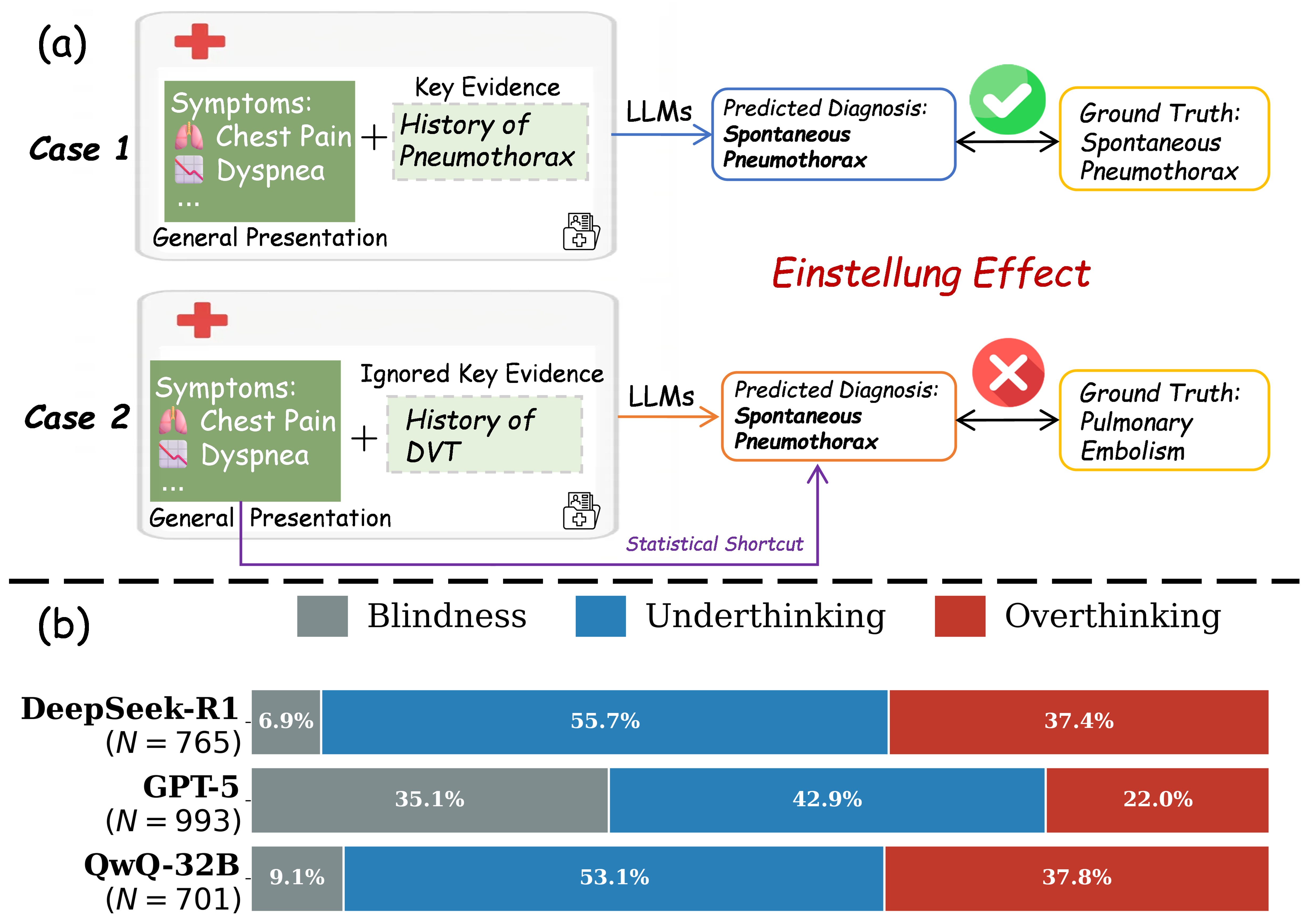} 
    \vspace{-0.3cm} 
    \caption{(a) Example of Einstellung Effect 
    (b) Distribution of failure modes under the Einstellung Effect across reasoning LLMs, including \textbf{Blindness} (missing key evidence), \textbf{Underthinking} (insufficient reasoning), and \textbf{Overthinking} (rationalizing incorrect priors).}
    \label{fig:failure_modes}
    \vspace{-0.4cm} 
\end{figure}

Although various medical benchmarks evaluate Med-LLMs\citep{medpalm,medbench2,medbench1}, they assess general medical capabilities rather than susceptibility to the Einstellung Effect. Existing benchmarks focus on knowledge evaluation (e.g., Medical QA on USMLE\citep{medqa,pal2022medmcqa}) or clinical task performance (e.g., Clinical Summarization\citep{johnson2023mimic} and Prognosis Prediction~\citep{nyutron,chen2024clinicalbench}), testing static knowledge recall and standardized procedures. The Einstellung Effect manifests critically in DDx scenarios requiring identification of subtle discriminative features between similar diseases. Detecting this effect requires a \textbf{counterfactual evaluation design}: presenting cases with similar symptoms but different diagnoses to assess whether models override pattern-based shortcuts for case-specific reasoning. However, current benchmarks lack such counterfactual scenarios. Thus, a specialized benchmark is needed to evaluate the Einstellung Effect in LLMs.

While current reasoning LLMs demonstrate strong logical capabilities, they remain susceptible to the Einstellung Effect in differential diagnosis. These models follow a "think-before-answer" paradigm but primarily establish simple symptom-disease associations rather than identifying discriminative evidence to disrupt pattern-based shortcuts. In Fig.~\ref{fig:failure_modes}, GPT-5 exhibits \textbf{blindness} in over 35\% of error cases—completely ignoring key discriminative symptoms and defaulting to stereotypical diagnoses. Among cases where key symptoms are acknowledged, 43\% involve \textbf{underthinking} (insufficient analysis) and 22\% involve \textbf{overthinking} (motivated reasoning). These patterns reveal that current models lack structured mechanisms for rigorous evidence analysis. In contrast, real-world clinical practice follows Evidence-Based Medicine (EBM)\citep{EBM} framework: (1) \textbf{Problem Representation}—objectively reconstructing patient conditions; (2) \textbf{Acquire \& Appraise}—actively seeking and verifying discriminative evidence; and (3) \textbf{Apply}—grounding diagnoses in verified evidence. Existing reasoning LLMs unfold reasoning linearly based on intuition, forcing a black-box "Symptoms → Diagnosis" mapping while neglecting the interpretable "\textbf{Symptoms → Evidence Verification → Diagnosis}" path. Therefore, constructing a reasoning framework grounded in EBM's cognitive architecture is imperative to mitigate the Einstellung Effect.

\textbf{\ourmethod:} To bridge these gaps, we introduce \textbf{\ourmethod}, a benchmark for evaluating the Einstellung Effect in medical LLMs via counterfactual differential diagnosis. \ourmethod~contains 5,383 paired clinical cases spanning 49 diseases across eight departments. To enable counterfactual evaluation, we employ a rigorous four-stage pipeline to generate the paired samples. Each pair consists of a \textit{control} case and a minimally edited \textit{trap} case: the trap case preserves most contextual evidence from the control case but replaces only the key discriminative evidences so that the correct diagnosis flips to a competing disease. This paired design creates counterfactual DDx scenarios in which superficial pattern matching strongly favors the original label, while correct diagnosis requires attending to the modified discriminative evidence. Using these pairs, we quantify susceptibility to the Einstellung Effect with \textbf{Bias Trap Rate}, the probability that a model—despite correctly solving the control case—misdiagnoses the trap case as the control label. We evaluate a broad set of 10 general and 5 medical-domain LLMs, as well as 2 LLM-based agents on \ourmethod, and observe substantial Einstellungs Effect errors across different models.

\textbf{ECR-Agent:} To mitigate the Einstellung Effect, we propose \textbf{ECR-Agent} (Evidence-based Causal Reasoning Agent), an agentic framework that emulates clinicians' EBM-grounded reasoning process through explicit discriminative evidence verification. ECR-Agent comprises two core components: (1) \textbf{Dynamic Causal Inference (DCI)} for structured diagnostic reasoning, and (2) \textbf{Critic-driven Graph and Memory Evolution (CGME)} for accumulating clinical experience.
The DCI module operationalizes the EBM framework through three stages. First, \textbf{dual-pathway perception} generates both intuitive differential diagnoses and an objective problem representation from patient symptoms, preventing premature diagnostic closure. Second, \textbf{dynamic causal graph reasoning} systematically seeks and verifies discriminative evidence through three progressive steps, each corresponding to a level in Pearl's causal hierarchy~\cite{pearl2018book}—moving from observing patterns to actively testing hypotheses to counterfactual verification: (i) \textbf{Causal Graph Initialization} (\textit{Association level}—observing correlations)—constructs a causal graph connecting observed symptoms, candidate diseases, and a pre-defined illness graph with prior illness knowledge to establish initial diagnostic hypotheses based on symptom-disease associations; (ii) \textbf{Forward Causal Reasoning} (\textit{Intervention level}—testing what happens if we seek new evidence)—actively retrieves discriminative evidence from external knowledge bases as pivot nodes while incorporating typical supporting evidence as general nodes, then evaluates how each piece of evidence supports or refutes competing diagnoses to prevent underthinking; (iii) \textbf{Backward Causal Reasoning} (\textit{Counterfactual level}—asking "what if this disease were true?")—performs counterfactual verification by identifying what evidence would be missing for each hypothesis, represented as shadow nodes that penalize incomplete diagnostic support and prevent overthinking. Third, the \textbf{evidence audit} module computes an evidence-based causal graph score for each candidate disease, generates graph summary with disease-centric subgraphs, retrieves similar cases from an exemplar base, and produces the final diagnosis grounded in verified evidence rather than pattern matching.
The CGME module enables experience accumulation across cases. Using a critic model, it iteratively refines diagnostic predictions until correctness is achieved, then stores: (1) case-level experience—the complete reasoning trace in the exemplar base for future retrieval; and (2) illness-level experience—merging and refining causal subgraphs across cases into consolidated illness graphs that capture refined discriminative patterns for each disease. Our contributions are as follows:
\begin{itemize}
    \item We propose \textbf{\ourmethod}, the first benchmark for evaluating the Einstellung Effect in medical LLMs, and introduce a novel metric revealing substantial model susceptibility.

    \item We propose \textbf{ECR-Agent}, an evidence-based framework to systematically verify discriminative evidence and accumulate clinical experience, mitigating the Einstellung Effect.
    
    \item Through extensive experiments, we demonstrate ECR-Agent's superiority and reveal current LLMs suffer from Einstellung Effect.
\end{itemize}


\section{Related Work}

\subsection{Medical LLMs and Agents}
LLMs have progressed from general medical assistants~\citep{medpalm,achiam2023gpt} passing USMLE exams to reasoning models using "think-before-answer" paradigms and LLM-based agents employing collaboration and retrieval. Agentic frameworks like MDAgents~\citep{kim2024mdagents} and MedAgents~\citep{medagents} use multi-role debate, while RAG systems like MedGraphRAG~\citep{medrag} and PrimeKG~\citep{primekg} incorporate Knowledge Graphs to reduce hallucinations.
However, current models suffer from the Einstellung Effect~\citep{alavi2023redherrings,kim2025limitations}, using associative "Symptoms → Diagnosis" mappings instead of systematically verifying discriminative evidence. This leads models to favor statistical shortcuts over patient-specific evidence, with multi-agent collaboration potentially amplifying Consensus Bias~\citep{biasmedqa}. We therefore introduce ECR-Agent, an Evidence-Based Medicine (EBM) agentic framework~\citep{EBM} that systematically verifies discriminative evidence through structured "Symptoms → Evidence Verification → Diagnosis" reasoning.

\subsection{Medical Benchmarks for LLMs}
Benchmarks for medical LLMs have shifted from static knowledge recall to dynamic reasoning. Early datasets like MedQA \citep{medqa} and PubMedQA \citep{pubmedqa} assess factual knowledge, while DDXPlus \citep{fansi2022ddxplus} and AgentClinic \citep{agentclinic} evaluate diagnostic processes. However, existing benchmarks typically employ Independent and Identically Distributed (I.I.D.) samples or standard clinical presentations. They lack adversarial and counterfactual designs required to expose the Einstellung Effect. High performance on these datasets may reflect statistical fitting rather than robust reasoning. Thus, we propose \ourmethod, a benchmark to evaluate the Einstellung Effect in medical LLMs via counterfactual differential diagnosis.

\section{\ourmethod~Benchmark}
\label{sec:benchmark}

\begin{figure*}[t!]
    \centering
    \includegraphics[width=\textwidth]{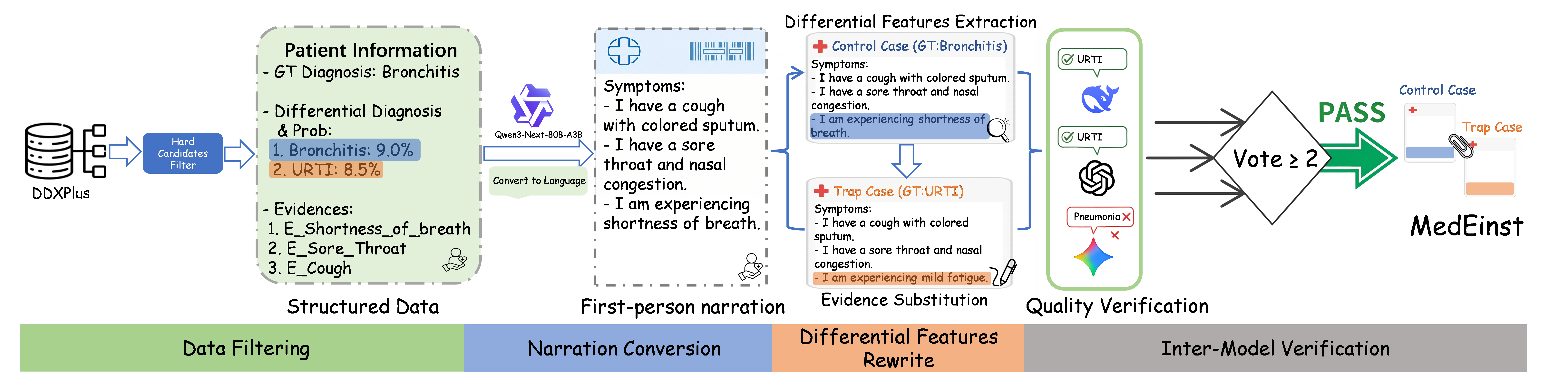}
    \caption{\textbf{Data construction of \ourmethod }
with four-stage process: (1) \textbf{Data Filtering} for hard candidates, (2) \textbf{Narration Conversion} to natural language, (3) \textbf{Differential Features Rewrite} for trap case generation, and (4) \textbf{Inter-Model Verification} for quality control.}
    \label{fig:MedEinst_pipeline}
\end{figure*}

\noindent\textbf{Overview.} We introduce \textbf{\ourmethod}, a benchmark to evaluate the Einstellung Effect in medical LLMs through counterfactual differential diagnosis via a four-stage construction pipeline (Fig.~\ref{fig:MedEinst_pipeline}). Moreover, we propose the \textbf{Bias Trap Rate} to quantify how often models solve a control case but fail a minimally edited trap case due to superficial reasoning.

\subsection{Problem Formulation}
We formalize medical diagnosis as a mapping $f: \mathcal{X} \to \mathcal{Y}$ ,where $\mathcal{X}$ denotes the patient narrative space and $\mathcal{Y}$ is the label space of 49 pathologies. We define a \textbf{Counterfactual Pair}($\mathbf{x}^c$,$\mathbf{x}^t$) consisting of: (1) \textbf{Control Case ($\mathbf{x}^c$)}, a typical presentation where statistical priors align with the ground truth (GT) $y_{gt}$; and (2) \textbf{Trap Case ($\mathbf{x}^t$)}, an adversarial variant generated via minimal modification. Crucially, $x^t$ remains statistically similar to $y_{gt}$ but logically implies a bias label $y_{bias}$ due to specific discriminative evidence.

\paragraph{Definition 1 (Einstellung Effect).} 
A model $f$ exhibits the Einstellung Effect if and only if:
\begin{equation}
    f(\mathbf{x}^c) = y_{gt} \quad \land \quad f(\mathbf{x}^t) = y_{gt}
\end{equation}
This implies that while the model demonstrates fundamental diagnostic competence (evidenced by success on the control case), it fails to rectify its prior intuition when confronted with the discriminative features in the trap case, rigidly persisting with the original diagnosis.

\subsection{Benchmark Construction}

\noindent\textbf{Data Filtering.} We collect 226,814 samples covering 49 pathologies from the DDXPlus dataset \citep{fansi2022ddxplus} $\mathcal{D}{src}$ and filter for "Hard Candidates" where evidence-based reasoning is strictly necessary. Specifically, we select samples where the probability gap between the ground truth diagnosis $y_{gt}$ and the top competing diagnosis $y_{bias}$ is less than 0.5\%, ensuring that prior probabilities alone cannot distinguish between diagnoses and forcing the model to perform evidence-based differential diagnosis.

\noindent\textbf{Narration Conversion.}
To simulate real-world clinical scenarios, we transform structured feature sets $\mathbf{s}$ into first-person natural language narratives $\mathbf{x}$ that capture the unstructured and noisy characteristics of actual medical records.

\noindent\textbf{Differential Features Rewrite.}
This module precisely induces the Einstellung trap while maintaining clinical validity. To prevent hallucination, we ground our generation in the DDXPlus Knowledge Base ($\mathcal{K}$) rather than using standard rewriting. Specifically, we first perform \textit{Differential Features Extraction} to identify the key discriminative features $k_{gt}$ that distinguish $y_{gt}$ from $y_{bias}$. Second, \textit{Trap Information} generation ($k_{trap}$) strictly derives misleading evidence from the bias disease knowledge base $K_{bias}$. Finally, \textit{Evidence Substitution} uses an LLM to replace $k_{gt}$ with $k_{trap}$, generating $x^t$. This ensures the trap case logically points to $y_{bias}$ while preserving all other contextual information from the control case.

\noindent\textbf{Inter-Model Verification.}
To ensure high-quality pairs, we employ an ``LLM-as-a-Judge'' committee $\mathcal{J} = \{\text{GPT-5}, \text{DeepSeek-R1}, \text{Gemini-2.5-Pro}\}$ to assess each pair $(x^c, x^t)$ across three dimensions: diagnostic correctness verifies whether $x^t$ logically points to $y_{bias}$, medical plausibility assesses alignment with real-world medical logic, and narrative fluency evaluates text coherence (See Appendix~\ref{app:qualityassurance} for details). A pair is included in MedEinst $\mathcal{S}_{final}$ only if at least two judges vote positively on diagnostic correctness. As shown in Appendix Fig.~\ref{fig:quality_metrics}, selected trap cases maintain high plausibility and fluency comparable to control cases, ensuring performance drops stem from reasoning failures rather than textual artifacts.

\subsection{Dataset Statistics}
\ourmethod~contains 5,383 counterfactual pairs of clinical narratives (10,766 cases total) covering 49 pathologies, derived from the DDXPlus test split to avoid data leakage. To provide an additional training set, we process and verify 10,689 pairs from the DDXPlus training split. 

\subsection{Quality Control}
To ensure clinical validity in \ourmethod, we implemented a rigorous quality control process involving four board-certified physicians with over 8 years of clinical experience. Our evaluation examined a stratified random sample of 1,500 counterfactual pairs (27.9\% of the dataset).
We developed a standardized scoring protocol evaluating seven binary quality dimensions: clinical plausibility of both control and trap cases, logical consistency of discriminative features, appropriateness of diagnoses, minimality of edits, and absence of artifactual patterns. Physicians evaluate each dimension through yes/no responses, with pairs satisfying all dimensions considered valid.
The quality assessment yielded strong results, with 96.1\% of evaluated pairs meeting our thresholds. Dimension-specific quality rates ranged from 94.3\% to 98.2\%. Inter-rater reliability analysis produced a Fleiss' kappa of 0.79, indicating substantial agreement. Pairs failing thresholds (3.9\%) were either revised (2.1\%) or excluded (1.8\%) to maintain benchmark integrity.

\subsection{Evaluation Metrics}
To quantify the Einstellung Effect, we first prompt the model to generate diagnostic results for all counterfactual pairs $(x^c, x^t)$. Then, we evaluate performance using three specific metrics based on the set of samples $S_{correct\_control}$ where the model correctly diagnosed the control case ($f(x^c) = y_{gt}$). \textbf{Baseline Accuracy} ($Acc_{base} = |S_{correct\_control}|/N_{total}$) establishes the model's fundamental diagnostic capability. \textbf{Robust Accuracy} ($Acc_{rob}= \sum_{i=1}^{N} \mathbb{I}(f(x^c_i) = y_{gt} \land f(x^t_i) = y_{bias})/N_{total}$) measures the proportion of pairs where the model correctly predicts both the control and trap cases. Finally, our primary metric, \textbf{Bias Trap Rate} ($R_{bias} = \sum_{i \in S_{correct\_control}} \mathbb{I}(f(x^t_i) = y_{gt})/|S_{correct\_control}|$), calculates the conditional probability that a capable model fall in the trap given that the model possesses the fundamental diagnostic capability. $N_{total}$ denotes the number of counterfactual pairs.

\begin{figure*}[t!]
    \centering
    \includegraphics[width=\textwidth]{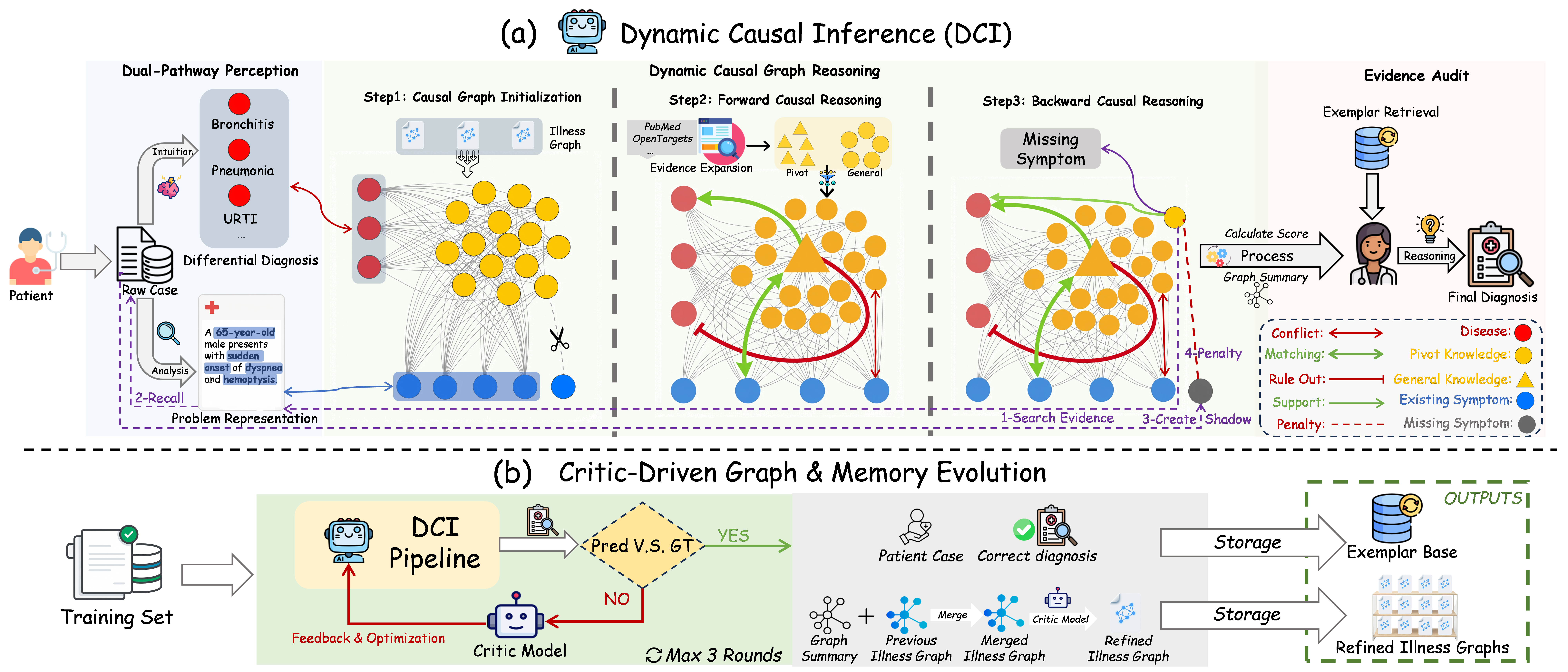}
    \caption{\textbf{ECR-Agent}, aligning LLM reasoning with Evidence-Based Medicine via two parts: (a) \textbf{Dynamic Causal Inference (DCI)} performs structured reasoning via \textit{dual-pathway perception}, \textit{dynamic causal graph reasoning} across three levels, and \textit{evidence audit} for final diagnosis. (b) \textbf{Critic-Driven Graph \& Memory Evolution (CGME)} iteratively refines the system by storing validated reasoning paths in an exemplar base and consolidating disease-specific knowledge into evolving illness graphs.}
    
    \label{fig:ecr_framework}
\end{figure*}

\section{ECR-Agent Framework}
\label{sec:agent}
\noindent\textbf{Overview.} To mitigate the Einstellung Effect, we propose ECR-Agent framework to align LLM reasoning with the rigorous verification standards of EBM (Fig.~\ref{fig:ecr_framework}). ECR-Agent comprises two synergistic components: (1) Dynamic Causal Inference (DCI), which performs structured diagnostic reasoning through dual-pathway perception, a three-level causal graph verification process (spanning association, intervention, and counterfactual levels), and evidence audit; and (2) Critic-driven Graph and Memory Evolution (CGME), which facilitates continuous improvement by refining diagnostic outputs and accumulating clinical experience into dynamic knowledge bases (Appendix Algorithm~\ref{alg:ecr_agent_full} and~\ref{implementation}).

\subsection{Critic-Driven Graph \& Memory Evolution}
To accumulate diagnostic experience, we execute the DCI pipeline on the training set $D_{train}$ and introduce a critic model $M_{critic}$ (GPT-5) to orchestrate iterative refinement (Fig.~\ref{fig:ecr_framework} (b)). For each training case where the base model's prediction diverges from GT label, $M_{critic}$ provides corrective feedback to optimize the reasoning path (maximum 3 rounds). Upon achieving correct diagnosis, the validated graph summary is merged with existing \textbf{illness graphs} $\mathcal{G} = \{G_y | y \in \mathcal{Y}\}$—a collection of disease-specific causal graphs initialized with the first graph summary—and further refined by the critic model to consolidate disease-level knowledge. Simultaneously, validated reasoning trajectories $(\mathbf{x}, y_{gt}, \text{Path})$ are stored in an \textbf{Exemplar Base ($\mathcal{M}$)} for case-based retrieval during inference.

\subsection{The Dynamic Causal Inference (DCI)}

\subsubsection{Dual-Pathway Perception}
To implement EBM's first principle of objective problem representation, we decouple statistical priors from factual observation through two parallel pathways. Firstly, the \textit{intuitive pathway} generates Top-$k$ candidate diagnoses $D_{set} = \{d_1, ..., d_k\}$ via Chain-of-Thought prompting, capturing pattern-based hypotheses. Secondly, the \textit{analytic pathway} produces a \textit{problem representation} that objectively summarizes key case features independent of diagnostic assumptions. From this representation, we extract structured patient observations $P_{obs} = \{p_1, ..., p_m\}$ and explicitly categorize each observation's status $s(p)$ as $\textit{Present}$ (affirmed), $\textit{Absent}$ (negated), or $\textit{Missing}$ (unmentioned). This dual-pathway design forces the model to acknowledge objective clinical facts before forming diagnostic conclusions, preventing premature closure driven by superficial pattern matching.


\subsubsection{Dynamic Causal Graph Reasoning (DCGR)}
DCGR aligns with Pearl's causal hierarchy through three levels: (1) \textbf{Causal Graph Initialization} (association) connects symptoms $P_{obs}$ with candidates $D_{set}$ via illness graphs $\mathcal{G}$; (2) \textbf{Forward Causal Reasoning} (intervention) retrieves and evaluates discriminative evidence; (3) \textbf{Backward Causal Reasoning} (counterfactual) penalizes hypotheses via expected-but-absent "shadow nodes". 


\noindent\textbf{Causal Graph Initialization.}
To establish initial diagnostic hypotheses based on observed correlations, we construct a causal graph integrating patient observations with disease knowledge. For each candidate $d \in D_{set}$, we retrieve its illness graph $G^{(d)}_{ill} = (V_d, V_p, V_k; E)$ from $\mathcal{G}$, where $V_d$, $V_p$, $V_k$ represent disease, symptom, and knowledge nodes, and $E$ denotes their relationships. We perform merge-or-prune based on embedding similarity between observations $P_{obs}$ and $V_p$, retaining relevant nodes and merging novel observations, yielding the contextualized initial graph $G_{ill}$.


\noindent\textbf{Forward Causal Reasoning.}
To prevent underthinking by actively seeking comprehensive discriminative evidence, we simulate the intervention: "What happens if we actively seek new evidence to differentiate among competing diseases?" We retrieve medical knowledge from external sources (PubMed, OpenTargets) and extract: (1) \textit{pivot nodes} $V_a$—discriminative evidence differentiating diseases; (2) \textit{general nodes} $V_b$—typical supporting evidence. We expand $G_{ill}$ with these nodes: $G'_{ill} = G_{ill} \cup (V_a, V_b)$. Using Qwen3-32B, we identify 5 causal relations: \textit{conflict}, \textit{matching}, \textit{rule out}, \textit{support}, and \textit{penalty}. For $V_p \leftrightarrow V_k$, we classify as conflict or matching; for $V_d \leftrightarrow V_k$, as rule out or support, producing the refined graph $G'_{ill}$.





\noindent\textbf{Backward Causal Reasoning.}
To prevent overthinking and motivated reasoning, we perform counterfactual verification asking: "If disease $d$ were true, what evidence should we observe?" For each $d$, we trace backward to identify supporting knowledge nodes $V_k^{(d)}$ and expected symptom nodes $V_p^{(d)}$. When knowledge node $v_k \in V_k^{(d)}$ lacks matching observations in $P_{obs}$, we trigger counterfactual verification (purple dashed line), re-examining the case text $x$. If evidence remains unverified, we instantiate a \textit{shadow node} $v_s$ (grey node) with a penalty edge to $d$, yielding a final causal graph $G^{\dagger}_{ill}$. Shadow nodes explicitly penalize hypotheses lacking expected evidence, ensuring diagnoses are grounded in verified evidence. 

\subsubsection{Evidence Audit}

\noindent\textbf{Graph Scoring and Summary}:
To quantify evidential support, we calculate an evidence-based causal graph score $S(d)$ for each candidate $d$: $S(d) = w_m N_{match}(d) - w_c N_{conf}(d) - w_s N_{shadow}(d)$, where $N_{match}(d)$, $N_{conf}(d)$, and $N_{shadow}(d)$ count edges with matching, conflict, and penalty relations, respectively, and $w_m, w_c, w_s$ are weighting hyperparameters. We then generate a \textit{Graph Summary} by reorganizing the causal graph $G^{\dagger}_{ill}$ into $k$ disease-centric subgraphs, each centered on a candidate diagnosis. This reorganization preserves all graph information while structuring evidence around each hypothesis to facilitate evidence auditing.

ECR-Agent then integrates three information streams to derive the final diagnosis $y^*$: (1) \textit{intuition}—initial reasoning from dual-pathway perception; (2) \textit{evidence}—graph summary and scores ${S(d)}$; (3) \textit{experience}—similar cases retrieved from exemplar base $\mathcal{M}$. This holistic audit ensures the diagnosis is grounded in verified evidence rather than pattern-based biases.


\begin{table*}[t]
\centering
\small 
\setlength{\tabcolsep}{12pt} 
\caption{
    Performance comparison of current LLMs and LLM-based Agents. 
}
\label{tab:main_results}
\begin{tabular}{l|c|ccc}
\toprule
\textbf{Model} & \textbf{Size} & \textbf{Baseline Acc} ($\uparrow$) & \textbf{Robust Acc} ($\uparrow$) & \textbf{Bias Trap Rate} ($\downarrow$) \\
\midrule
\multicolumn{5}{c}{\textit{\textbf{Open-Source LLMs}}} \\ 
\midrule
Kimi-k2 & - & 47.82 & 12.46 & 47.12 \\
DeepSeek-R1 & - & 42.20 & 11.32 & 46.12 \\
ZhipuAI/GLM-4.6 & - & 39.65 & 11.25 & 47.63 \\
Qwen/Qwen3-14B & 14B & 44.12 & 11.28 & 54.19 \\
Qwen/QwQ-32B & 32B & 41.05 & 11.14 & 44.88 \\
Qwen/Qwen3-32B & 32B & 40.25 & 11.86 & 43.46 \\
Qwen/Qwen3-235B-A22B & 235B & 40.51 & 11.40 & 43.32 \\
\midrule
\multicolumn{5}{c}{\textit{\textbf{Proprietary LLMs}}} \\ 
\midrule
GPT-5 & - & \underline{54.30} & \underline{15.78} & 51.87 \\
Gemini-2.5-pro & - & 53.58 & 10.97 & 60.90 \\
Claude-Sonnet-4.5 & - & 42.09 & 12.36 & 42.98 \\ 
\midrule
\multicolumn{5}{c}{\textit{\textbf{Medical-Specific LLMs}}} \\
\midrule
Lingshu-7B & 7B & 14.93 & 3.20 & \underline{36.74} \\
Llama3-Med42-8B & 8B & 6.51 & 1.19 & 44.00 \\
MedGemma-27B-text-it & 27B & 40.92 & 11.68 & 53.88 \\ 
Baichuan-M2-32B & 32B & 45.03 & 7.18 & 66.10 \\
Lingshu-32B & 32B & 27.68 & 6.17 & 54.20 \\
\midrule
\multicolumn{5}{c}{\textit{\textbf{LLM-based Agents}}} \\ 
\midrule
MDAgent (Qwen3-32B) & 32B & 29.70 & 10.34 & 40.34 \\
DyLAN (Qwen3-32B) & 32B & 32.11 & 8.11 & 41.69 \\
\textbf{ECR-Agent (Qwen3-32B)} & \textbf{32B} & \textbf{69.49} & \textbf{24.21} & \textbf{33.75} \\ 
\bottomrule
\end{tabular}
\end{table*}

\section{Experiments}
\label{sec:experiments}

\subsection{Evaluation Baselines}
We compare ECR-Agent against 3 baseline types: 1) \textbf{General LLMs:} state-of-the-art proprietary (GPT-5, Claude-Sonnet-4.5, Gemini-2.5-Pro) and open-source LLMs (DeepSeek-R1, Qwen3-32B, QwQ-32B); 2) \textbf{Medical LLMs:} Lingshu-7B, Llama3-Med42-8B, MedGemma-27B-text-it, Baichuan-M2-32B and Med42-8B; 3) \textbf{LLM-based Agent:} {MDAgent}~\citep{kim2024mdagents} and DyLAN~\citep{dylan}.

\begin{figure*}[t]
    \centering
    \includegraphics[width=\linewidth]{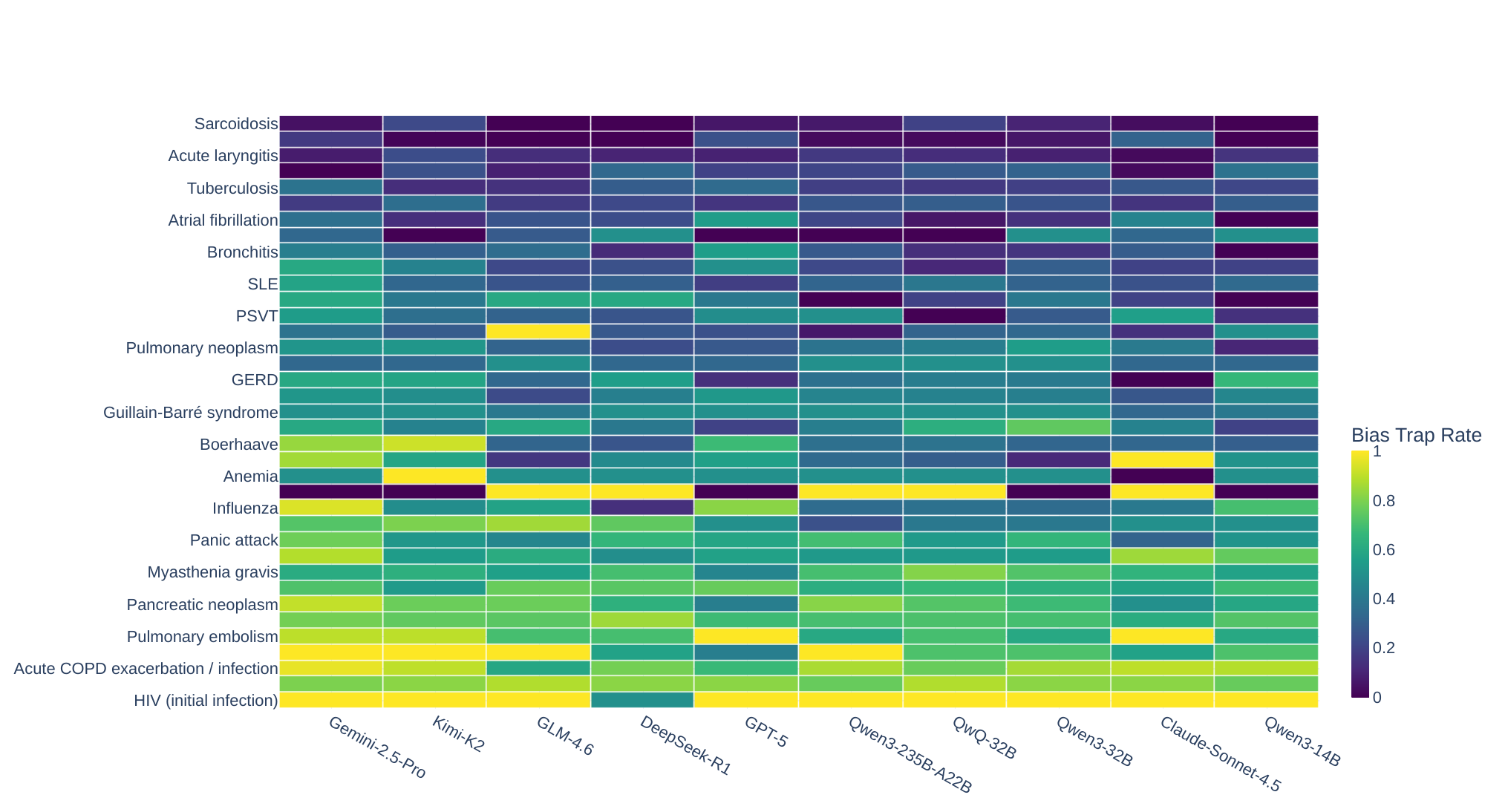} 
    \caption{Bias Trap Rate heatmap across diseases. The clustering indicates that models learn spurious correlations for common diseases (e.g., Pneumonia), leading to consistent bias.}
    \label{fig:heatmap}
\end{figure*}
\subsection{Overall Performance Comparison}
Table~\ref{tab:main_results} reveals a striking gap between diagnostic capability and robustness. While frontier models like GPT-5 and Gemini-2.5-Pro achieve the highest baseline accuracy (54.30\% and 53.58\%), they exhibit disproportionately high Bias Trap Rates ($>$50\%), indicating a fundamental trade-off where models that better fit general medical distributions develop stronger priors that aggressively filter out low-probability counter-evidence (Perceptual Blindness, Fig.~\ref{fig:failure_modes}), making them more susceptible to Einstellung traps than weaker models. Agent frameworks like MDAgent (multi-role debate) and DyLAN (dynamic agent selection) show low robust accuracy (\textasciitilde8-10\%) and high trap rates due to noise amplification, where dynamic interaction topologies merely reinforce the dominant statistical prior (Consensus Bias) rather than correcting it—DyLAN's strategy of selecting "high-contribution" agents exacerbates this by favoring agents that align with the incorrect group consensus. In contrast, ECR-Agent achieves substantial improvements (69.49\% baseline accuracy, 24.21\% robust accuracy, 33.75\% bias trap rate), empirically validating that resolving the Einstellung Effect requires a paradigm shift from statistical fitting (probability) to causal verification (evidence).



\subsection{{Ablation Study}}
We conduct an ablation study on ECR-Agent (Qwen3-32B as the base model) to evaluate the module effectiveness. In Table~\ref{tab:ablation}, adding DCI substantially improves Base Accuracy from 40.25\% to 55.49\%, showing the effectiveness of structured causal reasoning. Further incorporating CGME yields additional significant gains to 69.49\% Base Accuracy and reduces Trap Rate to 33.75\%, proving the critical role of experience accumulation. 


\begin{table}[t]
    \centering
    \small 
    \setlength{\tabcolsep}{3.5pt}
    \caption{\textbf{Ablation Study on ECR-Agent components.}} 
    \label{tab:ablation}
    \resizebox{\linewidth}{!}{
    \begin{tabular}{cc|ccc}
        \toprule
        \textbf{DCI} & \textbf{CGME} & \textbf{Base Acc} ($\uparrow$) & \textbf{Rob Acc} ($\uparrow$) & \textbf{Trap Rate} ($\downarrow$) \\
        \midrule
        & & 40.25 & 11.86 & 43.46 \\
        \checkmark &  & 55.49 & 19.94 & 38.32 \\
        \checkmark & \checkmark & \textbf{69.49} & \textbf{24.21} & \textbf{33.75} \\
        \bottomrule
    \end{tabular}
    }
\end{table}

\subsection{Disease-Specific Analysis}
Fig.~\ref{fig:heatmap} reveals heterogeneity in Bias Trap Rates across diseases, exposing the structural nature of the Einstellung Effect: a systematic ``High-Bias Cluster'' emerges in diseases like \textit{Pulmonary Embolism} and \textit{Initial HIV Infection} (bottom rows) whose presentations overlap with high-prevalence distractors (e.g., Flu, Anxiety), where LLMs learn spurious correlations between generic symptoms and statistically probable diagnoses while ignoring key discriminative evidence. This failure persists across all architectures, e.g. reasoning-optimized (DeepSeek-R1) and massive-scale LLMs (Qwen3-235B), showing CoT capabilities as pattern matchers that collapse when diagnosis requires overriding priors with specific evidence. 

\begin{figure}[t]
    \centering
    \includegraphics[width=\linewidth]{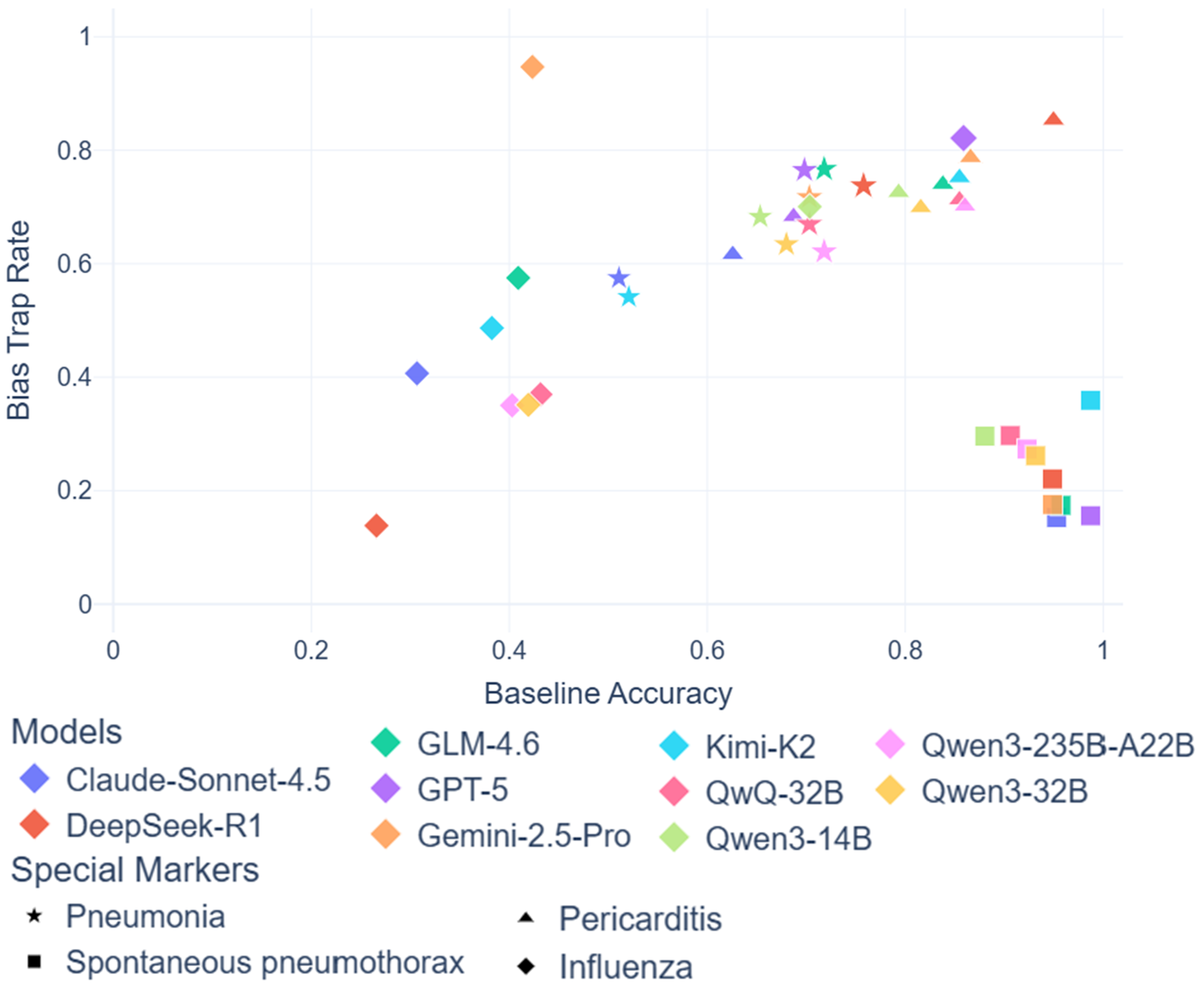} 
    \caption{Baseline Accuracy vs. Bias Trap Rate. }
    \label{fig:scaling}
\end{figure}

\subsection{Scaling Laws vs. Einstellung Effect}
Fig.~\ref{fig:scaling} reveals a failure of Scaling Laws in robust medical reasoning: no meaningful correlation exists between model size and robustness, with frontier models like GPT-5 and Gemini-2.5-Pro occupying a ``High Capability, High Bias'' region where scaling improves baseline diagnostic capability ($ACC_{base}$) but paradoxically exacerbates Einstellung susceptibility. We term this \textbf{``Stronger Priors, Stronger Blindness''}: larger models capture statistical regularities so effectively they become overconfident in initial intuitions, making it harder to override diagnoses when presented with subtle counter-evidence—a trend evident across model tiers (e.g., Gemini-2.5-Pro achieves superior baseline accuracy yet exhibits a 60.90\% bias trap rate, significantly higher than less capable models). These findings demonstrate the Einstellung Effect as a fundamental cognitive failure mode that persists with scale, necessitating architectural interventions like ECR-Agent that decouple evidence verification from probabilistic generation.

\section{Conclusion}
We introduced \ourmethod, the first counterfactual benchmark exposing the Einstellung Effect in medical LLMs, revealing that frontier models achieve high baseline accuracy yet remain severely susceptible to statistical shortcuts. We proposed ECR-Agent, which aligns LLM reasoning with Evidence-Based Medicine through structured causal inference and knowledge evolution. 

\section*{Limitations}
While MedEinst includes 5,383 counterfactual pairs, it currently covers only 49 common pathologies across eight departments. Although these diseases represent high-frequency diagnostic scenarios in emergency medicine, they constitute a small fraction of the vast medical ontology (e.g., ICD-10). Consequently, the manifestation of the Einstellung Effect in rare diseases or complex comorbidities remains to be fully explored. We view MedEinst as a foundational proof-of-concept, paving the way for future benchmarks to expand into broader disease taxonomies.

\bibliography{custom}

\appendix
\clearpage
\raggedbottom
\textbf{\Large Appendix}
\setcounter{table}{0}
\renewcommand{\thetable}{A\arabic{table}}

\textbf{Abstract.} This appendix provides supplementary materials for the MedEinst benchmark and the ECR-Agent framework.

    
    \textbf{Appendix A} details the methodological algorithms for benchmark construction and agent inference, along with the causal graph schema and evaluation metrics.
    
    \textbf{Appendix B} provides a comprehensive analysis of the MedEinst benchmark, including clinical specialty distribution, quality assurance protocols, and dataset statistics.
    
    \textbf{Appendix C} outlines the implementation details, including experimental settings and baseline configurations.
    
    \textbf{Appendix D} presents additional empirical analyses, focusing on detailed failure modes and the capability-robustness gap.
    
    \textbf{Appendix E} offers a concrete case study (Case 100473) to qualitatively demonstrate the reasoning trace and interpretability of our approach.
    
     \textbf{Appendix F} extends the discussion on theoretical grounding, mapping our framework to the Causal Hierarchy and contrasting it with existing paradigms.
    
    \textbf{Appendix G} displays raw data samples illustrating the input format.
    
    \textbf{Appendix H} lists the detailed prompts used for data construction and the agent reasoning pipeline.

\vspace{10pt}
\section{Methodological Details}
\label{app:method}

\subsection{MedEinst Construction Algorithm}
Algorithm \ref{alg:medeinst_pipeline} outlines the rigorous four-stage pipeline employed to construct the MedEinst benchmark. The process begins with Data Filtering to select "Hard Candidates" where statistical shortcuts fail. It then proceeds to Narration Conversion and Differential Features Rewrite, transforming structured data into natural language and injecting adversarial traps based on knowledge base from DDXPlus \citep{fansi2022ddxplus}. Finally, Inter-Model Verification serves as a quality control filter, ensuring that the generated trap cases are medically plausible.

\renewcommand{\algorithmicrequire}{\textbf{Input:}}
\renewcommand{\algorithmicensure}{\textbf{Output:}}
\begin{algorithm}[H]
\footnotesize
\caption{Construction Pipeline of \textbf{MedEinst}}
\label{alg:medeinst_pipeline}
\begin{algorithmic}[1] 
    \Require Source dataset $\mathcal{D}_{src}$, Knowledge Base $\mathcal{K}$ (DDXPlus), LLM Judge Committee $\mathcal{J}$; Threshold $\epsilon = 0.5\%$.
    \Ensure Paired Counterfactual Benchmark $\mathcal{S}_{final}$.

    \State Initialize $\mathcal{S}_{final} \leftarrow \emptyset$
    \For{each sample $(\mathbf{s}, y_{gt}, P) \in \mathcal{D}_{src}$}
        
        \State \textbf{Step 1: Data Filtering}
        \If{$|P(y_{gt}) - P(y_{bias})| < \epsilon$}
            
            \State \textbf{Step 2: Narration Conversion}
            \State $x^c \leftarrow \text{LLM}(\mathbf{s})$ 
            
            \State \textbf{Step 3: Differential Features Rewrite}
            \State Retrieve $K_{gt}, K_{bias} \leftarrow \text{Query}(\mathcal{K}, \{y_{gt}, y_{bias}\})$
            \State $k_{gt} \leftarrow \text{LLM}(x^c, K_{gt}, K_{bias})$ 
            \State $k_{trap} \leftarrow \text{LLM}(K_{bias}, k_{gt})$ 
            \State $x^t \leftarrow \text{LLM}(x^c, k_{trap}, k_{gt})$ 
            
            \State \textbf{Step 4: Inter-Model Verification}
            \State $V_{score} \leftarrow \sum_{j \in \mathcal{J}} \mathbb{I}( \text{LLM}_j(x^t, y_{bias}) = \text{Correct})$
            
            \If{$V_{score} \geq 2$} 
                \State $\mathcal{S}_{final} \leftarrow \mathcal{S}_{final} \cup \{(x^c, x^t, y_{gt}, y_{bias})\}$
            \EndIf
        \EndIf
    \EndFor
    \State \Return $\mathcal{S}_{final}$
\end{algorithmic}
\end{algorithm}

\subsection{ECR-Agent Inference Algorithm}
Algorithm \ref{alg:ecr_agent_full} formally describes the complete workflow of the ECR-Agent, integrating both the training and inference phases. The algorithm first details the Critic-Driven Graph \& Memory Evolution (CGME), where the system iteratively refines illness graphs and accumulates an exemplar base using critic feedback on the training set. Subsequently, it presents the Dynamic Causal Inference (DCI) pipeline used during inference, which orchestrates Dual-Pathway Perception, Dynamic Causal Graph Reasoning (across initialization, forward, and backward steps), and the final Evidence Audit to derive robust diagnoses for unseen cases.

\subsection{Causal Reasoning Graph}
\textbf{Graph Schema Definition}:
\begin{itemize}
    \item \textbf{Patient Nodes($V_P$)}: Encode structured clinical observations extracted from problem representation. Crucially, we distinguish node status $s(p)$ into three states: $\textit{Present}$ (affirmed), $\textit{Absent}$ (negated), $\textit{Missing}$ (unmentioned).
    \item \textbf{Knowledge Nodes($V_K$)}: Encode disease-specific clinical entities (e.g., symptoms, biomarkers) distilled from literature. They are categorized into \textbf{General} (typical features) and \textbf{Pivot} (discriminators).
\end{itemize}

\textbf{Merge-or-Prune operation}
\small
\begin{equation}
\text{Action}(p_{script}) = 
\begin{cases} 
\text{Merge}, & \text{if }\cos(\mathbf{e}_{p_{script}}, \mathbf{e}_{p_{obs}}) > \tau \\
\text{Prune}, & \text{otherwise}
\end{cases}
\end{equation}
where $\tau=0.9$. This ensures $G_{curr}$ only contains the patient's actual data while inheriting relevant causal structures from the illness graphs.

\begin{algorithm}[H]
\footnotesize
\caption{ECR-Agent Evolution \& Inference Pipeline}
\label{alg:ecr_agent_full}

\begin{algorithmic}[1] 
    \Require Training Set $\mathcal{D}_{train}$; New Case $\mathbf{x}_{new}$
    \Ensure Refined Illness Graphs $\mathcal{G}_{refined}$; Exemplar Base $\mathcal{M}$; Diagnosis $d^\star$; Causal Reasoning Graph $G_{ill}^{(d^\star)}$

    \Statex 
    \State \textbf{/* Critic-Driven Graph \& Memory Evolution */}
    \For{sample $(\mathbf{x}, y_{gt}) \in \mathcal{D}_{train}$}
        \State $t \leftarrow 0$
        \While{$t < 3$}
            \State $(d_{pred}, G_{summary}) \leftarrow \text{DCI\_Pipeline}(\mathbf{x})$
            
            \If{$d_{pred} == y_{gt}$}
                \State Load Previous Graph $G_{prev}$ for $y_{gt}$
                \State $G_{merged} \leftarrow \text{Merge}(G_{prev}, G_{summary})$
                \State \textbf{break}
            \Else
                \State $\text{ApplyCriticFeedback}(\mathbf{x})$
                \State $t \leftarrow t + 1$
            \EndIf
        \EndWhile
        \State \Comment{If loop ends without success, sample is discarded.}
    \EndFor

    \Statex
    \State \textbf{/* Dynamic Causal Inference (DCI) Pipeline */}
    \Function{DCI\_Pipeline}{$\mathbf{x}$}
        \State \textbf{Dual-Pathway Perception}
        \State $D_{set} \gets \text{IntuitivePathway}(x)$ 
        \State $P_{obs} \gets \text{AnalyticPathway}(x)$ 
        
        \State \textbf{Dynamic Causal Graph Reasoning}
        \For{candidate $d \in D_{top}$}
            \State Load $G_{ill}^{(d)} = (V_p, V_k, E)$
            \State \textit{Step 1: Causal Graph Initialization}
            \State $V_{init} \leftarrow \{ p \in V_p \mid \text{Sim}(p, P_{obs}) > \tau \}$
            \State $G_{ill} \leftarrow \text{Initialize}(V_{init}, E)$
            
            \State \textit{Step 2: Forward Causal Reasoning}
            \State $V_{k} \leftarrow \text{LiveSearch}(d)$ 
            \State \Comment{Expand Pivot/General Nodes}
            \State $G^{'}_{ill} \leftarrow G_{ill} \cup \text{Link}(V_{d}, V_{k}) \cup \text{Link}(V_{k},V_{p})$
            
            \State \textit{Step 3: Backward Causal Reasoning}
            \State $\Delta_{miss} \leftarrow \{ k \in V^{(d)}_{k} \mid k \notin P_{obs} \land \text{IsExpected}(k) \}$
            \State $N_{shadow}^{(d)} \leftarrow \emptyset$

            \For{$k \in \Delta_{miss}$} 
                \If{$\text{ReExamine}(\mathbf{x}, k) == \text{Found}$}
                    \State $P_{obs} \leftarrow P_{obs} \cup \{k\}$; Update $G^{\dagger}_{ill}$
                \Else
                    \State $N_{shadow}^{(d)} \leftarrow N_{shadow}^{(d)} \cup \{k\}$ 
                    \State \Comment{Create Shadow Node}
                \EndIf
            \EndFor
            
            \State \Comment{This $G^{\dagger}_{ill}$ serves as the "Graph Summary"}
            \State $Score(d) \leftarrow \text{CalculateScore}(G^{\dagger}_{ill}, N_{shadow}^{(d)})$
        \EndFor
        
        \State \textbf{Evidence Audit}
        \State $\mathcal{M}_{sim} \leftarrow \text{RetrieveExemplars}(\mathcal{M}, P_{obs})$
        \State $d^\star \leftarrow \text{LLM\_Judge}(D_{top}, \{Score(d)\}, \mathcal{M}_{sim})$
        \State \Return $(d^\star, G_{ill}^{(d^\star)})$
    \EndFunction
\end{algorithmic}
\end{algorithm}

\subsection{Evaluation Metrics}

To precisely quantify the Einstellung Effect, we classify the model's predictions on paired samples ($x^c, x^t$) into three categories based on the intersection of their outcomes. Let $S_{correct\_control}$ denote the set of samples where the model correctly diagnoses the Control Case ($f(x^c) = y_{gt}$). We define the following metrics:

\begin{itemize}
    \item \textbf{Baseline Accuracy ($Acc_{base}$):} Measures the fundamental diagnostic capability on standard clinical presentations.
    \begin{equation}
        Acc_{base} = \frac{|S_{correct\_control}|}{N_{total}}
    \end{equation}
    
    \item \textbf{Robust Accuracy ($Acc_{rob}$):} Measures the proportion of pairs where the model maintains correctness across both control and trap cases (Robust Success).
    \begin{equation}
        Acc_{rob} = \frac{\sum_{i=1}^{N} \mathbb{I}(f(x^c_i) = y_{gt} \land f(x^t_i) = y_{bias})}{N_{total}}
    \end{equation}
    
    \item \textbf{Bias Trap Rate ($R_{bias}$):} The core metric for the Einstellung Effect. It measures the conditional probability of fall in the trap given that the model possesses the fundamental diagnostic capability. 
    \begin{equation}
        R_{bias} = \frac{\sum_{i \in S_{correct\_control}} \mathbb{I}(f(x^t_i) = y_{gt})}{|S_{correct\_control}|}
    \end{equation}
\end{itemize}

\section{MedEinst Benchmark Details}
\label{app:dataset}

\subsection{Clinical Specialty Analysis}
To assess the clinical breadth and diversity of the \textbf{MedEinst Benchmark}, we categorized the 49 target pathologies into 10 distinct clinical specialties. Unlike rigid anatomical classifications (e.g., ICD-10), we adopted a clinical taxonomy based on medical specialties and triage departments. This approach better reflects real-world diagnostic workflows where pathologies presenting with overlapping symptoms are managed by specific domains. 

\begin{figure}[htbp]
    \centering
    \includegraphics[width=\linewidth]{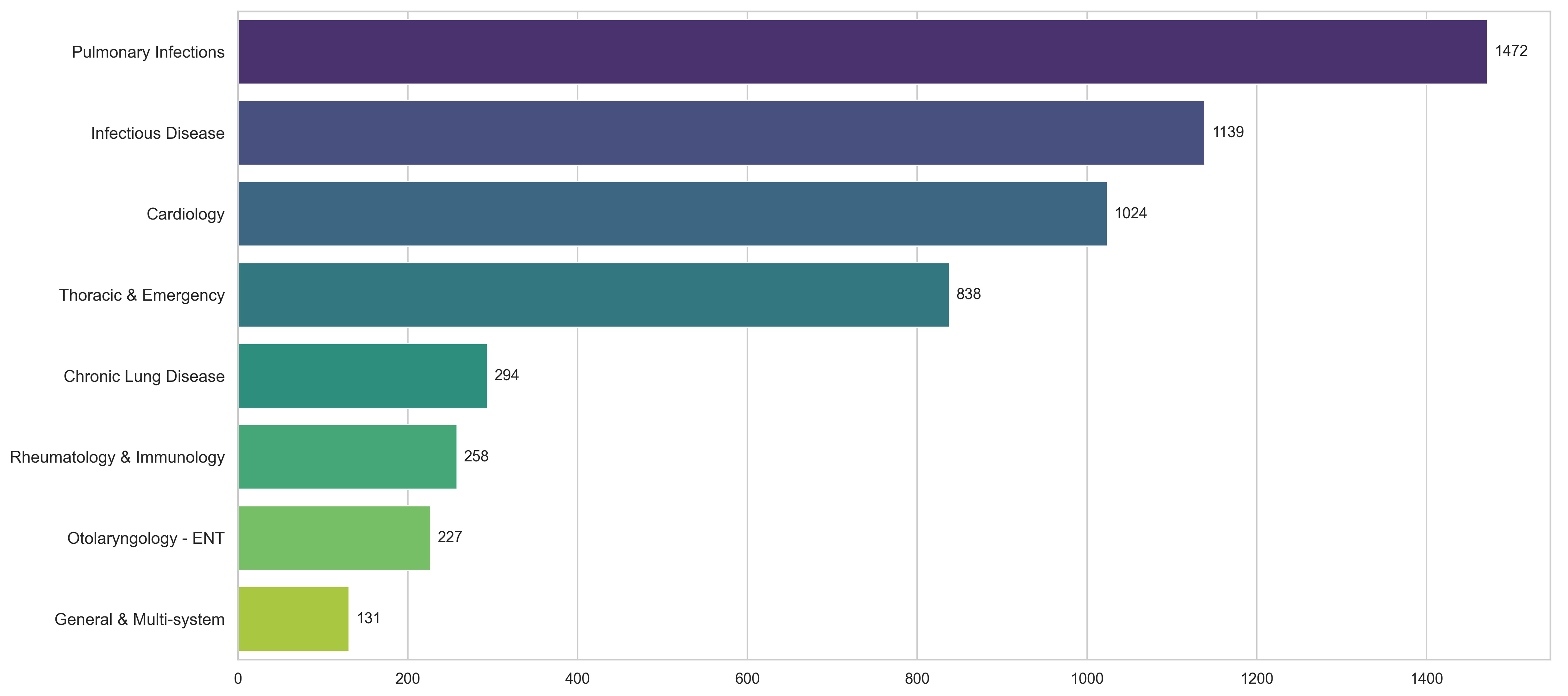}
    \caption{\textbf{Distribution of MedEinst Benchmark Pairs by Clinical Specialty.} The 5,383 test pairs are grouped into 10 categories based on standard clinical taxonomy. The high representation of Pulmonary and Cardiology cases reflects the dataset's focus on acute care scenarios where differential diagnosis is most critical.}
    \label{fig:specialty_dist}
\end{figure}

\subsection{Quality Assurance}
\label{app:qualityassurance}
To verify that our \textbf{Differential Features Rewrite} (Method \S 3.1.2) does not degrade the linguistic or clinical quality of the patient narratives, we analyzed the distribution of \textit{Medical Plausibility} and \textit{Narrative Fluency} scores assigned by the judge committee $\mathcal{J} = \{\text{GPT-5}, \text{DeepSeek-R1}, \text{Gemini-2.5-Pro}\}$.

\begin{figure}[htbp]
    \centering
    \includegraphics[width=0.95\linewidth]{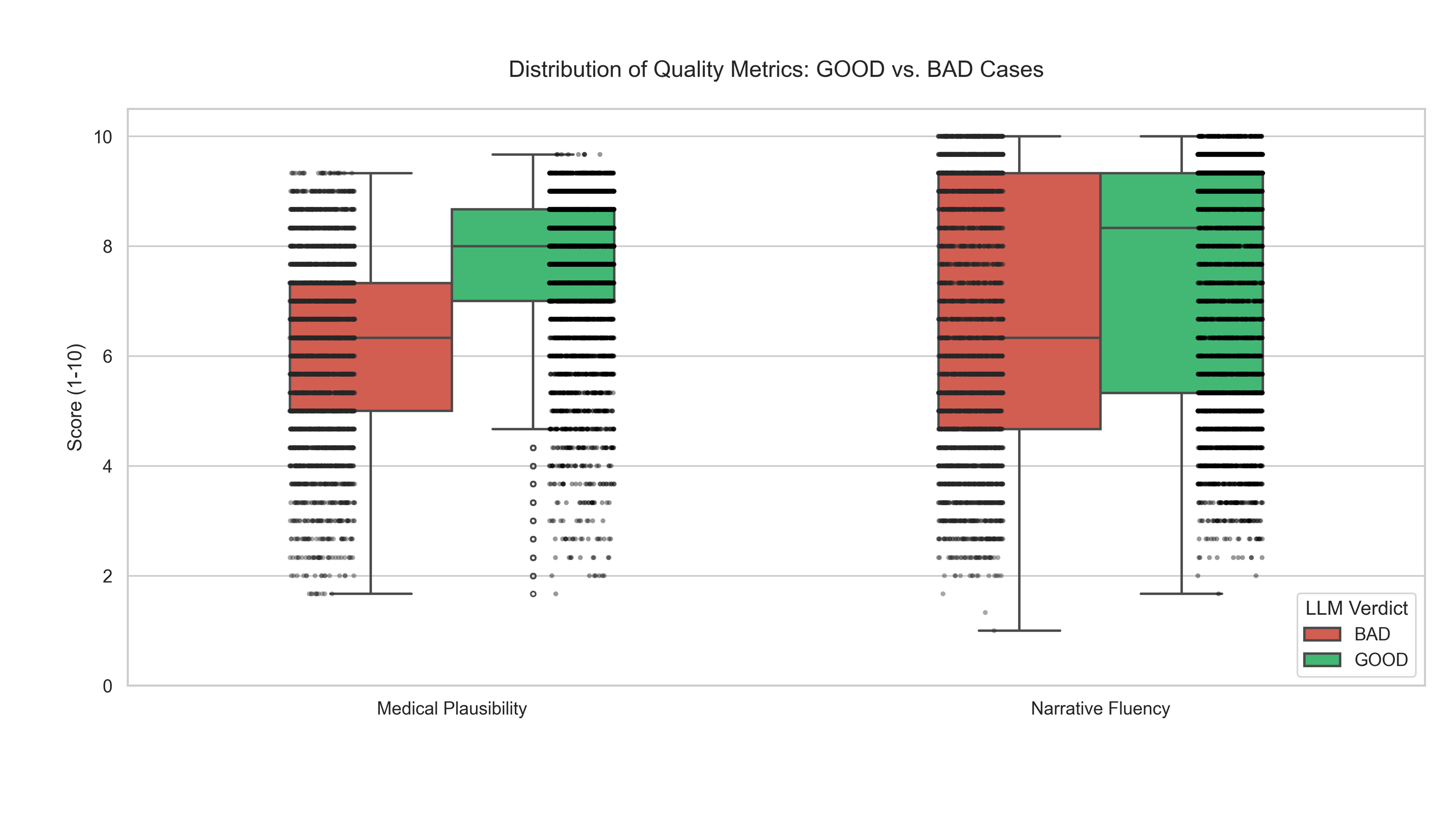}
    \caption{Distribution of quality metrics (Medical Plausibility and Narrative Fluency) for accepted (GOOD) versus rejected (BAD) trap cases. The high scores of accepted cases validate the effectiveness of our isomorphic rewriting protocol.}
    \label{fig:quality_metrics}
\end{figure}

Figure~\ref{fig:quality_metrics} presents the comparative analysis between \textbf{GOOD Cases} (successfully generated traps that passed verification) and \textbf{BAD Cases} (rejected traps).

\begin{itemize}
    \item \textbf{Medical Plausibility}: The GOOD cases (green boxplots) maintain a high median score ($\approx 8.0/10$), statistically indistinguishable from the original clinical notes. This confirms that the injected \textit{trap\_info} aligns logically with the patient's context (e.g., age, gender, symptoms, antecedents).
    \item \textbf{Narrative Fluency}: The rewriting process preserves the natural flow of the text, with GOOD cases achieving a median fluency score of $\approx 8.3/10$. In contrast, BAD cases often exhibit disjointed insertions or grammatical inconsistencies, justifying their exclusion.
\end{itemize}

This quality audit confirms that the \textbf{Einstellung Effect} observed in our benchmark stems from the model's inability to process conflicting evidence, rather than poor data quality.

\subsection{Dataset Statistics}
We constructed \textbf{MedEinst} based on the DDXPlus dataset, strictly adhering to its original chronological split to prevent data leakage. The benchmark comprises two subsets:
\begin{itemize}
    \item \textbf{Test Set (The Benchmark):} Derived from the DDXPlus test split, this set contains 5,383 counterfactual pairs of clinical narratives (totaling 10,766 cases) covering 49 pathologies. A unique feature of MedEinst is its \textit{Paired Counterfactual} design: each Control Case ($x^c$) is paired with a Trap Case ($x^t$) that differs only in \textbf{Key Discriminative Features}, yet leads to a contradictory diagnosis ($y_{gt}$ vs. $y_{bias}$). This design strictly decouples a model's statistical intuition from its logical reasoning capability.
    \item \textbf{Reference Set (Training Resource):} Derived from the DDXPlus training split, we processed and verified 10,689 pairs. This large-scale set is provided to support various research paradigms, including fine-tuning, few-shot learning, or RAG-based retrieval.
\end{itemize}

\noindent \textbf{Selection Criteria.} A sample pair is included in the final MedEinst benchmark $\mathcal{S}_{final}$ if and only if it receives a positive vote on Diagnostic Correctness from at least two judges. As shown in Appendix A, the selected trap cases maintain high medical plausibility and narrative fluency comparable to control cases. This rigorous verification ensures that performance drops in MedEinst stem from reasoning failures (Einstellung Effect) rather than textual artifacts or data noise.




\section{Implementation Details}
\label{implementation}
To simulate a realistic clinical diagnosis scenario where physicians encounter unseen cases, all baseline models and agent frameworks operate under a Zero-shot Chain-of-Thought (CoT) setting. 
For our \textbf{ECR-Agent}, we maintain the same zero-shot input for fair comparison. 
Specifically, in the \textit{Dual-Pathway Perception} phase, we configure the agent to generate the Top-$k$ candidate diagnoses with \boldmath$k=5$. This threshold was empirically selected to ensure sufficient coverage of potential differentials (including the ground truth and trap) while maintaining computational efficiency for the subsequent causal graph construction.
Evidence Expansion is supported by structured queries to OpenTargets and PubMed APIs, functioning as an extension of the agent's analytic system.

We evaluate all methods on the MedEinst benchmark (5,383 pairs). To drive the Critic-Driven Graph \& Memory Evolution, we utilized the MedEinst-Support set. To demonstrate the data efficiency of our framework, we did not employ the full support set. Instead, we curated a compact Balanced Seed Subset consisting of only 853 cases (approximately 8\% of the available training data).This subset was constructed using a Capped Sampling Strategy: we randomly sampled a maximum of $N=20$ cases per pathology, while retaining all available samples for rare diseases. This lightweight selection ensures that the agent can initialize robust \textit{Illness Graphs} and the \textit{Exemplar Base} with minimal data consumption, highlighting the framework's capability to generalize from sparse but balanced clinical examples.

\section{Additional Experimental Analysis}
\label{app:analysis}
To investigate the microscopic mechanisms and macroscopic characteristics of the Einstellung Effect, we conduct a multi-dimensional empirical analysis.

\subsection{Detailed Failure Mode Analysis}
To understand the cognitive failures behind Einstellung Traps, we conducted a fine-grained failure analysis on three representative models (DeepSeek-R1, GPT-5, QwQ-32B). We classify reasoning failures into three modes based on the model's interaction with the Key Discriminative Evidence. The classification was performed by a GPT-5 Auditor and \textbf{verified by human experts on a subset of data} (Cohen's $\kappa > 0.8$).

As shown in Figure~\ref{fig:failure_modes}, the distribution reveals distinct cognitive deficits:
\begin{itemize}
    \item \textbf{Blindness} Models completely fail to mention the key evidence in their CoT. This suggests that strong statistical priors filter out "unexpected" symptoms during the initial perception stage. \textit{Our Solution (Dual-Pathway Perception):} We introduce \textbf{Dual-Track Perception}, forcing the explicit extraction of a structured Problem Representation to ensure all evidence is "seen".
    \item \textbf{Underthinking} Even when evidence is seen, models often default to the most likely candidate without rigorous falsification. \textit{Our Solution (Causal Graph Reasoning):} We implement \textbf{Causal Graph Reasoning}. By constructing a patient-specific graph with Pivot Nodes, we structurally force bidirectional reasoning (Forward Support \& Backward Exclusion) to prevent the dismissal of contradictory evidence.
    \item \textbf{Overthinking} Advanced models (e.g., GPT-5) engage in \textbf{Motivated Reasoning}, hallucinating mechanisms to force-fit contradictions into the incorrect diagnosis. \textit{Our Solution (Evidence Audit):} We deploy an \textbf{Evidence Audit}. By performing Counterfactual Checks, the agent detects and penalizes such non-causal rationalizations, breaking the self-confirming loop.
\end{itemize}

\subsection{Overall Performance Comparison}
Table~\ref{tab:main_results} presents the performance of various models and agent frameworks on MedEinst. We observe three critical phenomena:

\paragraph{1. The Capability-Robustness Gap.}
While frontier models like GPT-5 and Gemini-2.5-Pro demonstrate superior fundamental diagnostic capabilities ($Acc_{base}$ of 54.30\% and 53.58\% respectively), their robustness remains disproportionately low, with $Acc_{rob}$ hovering around 10\%--15\%. Alarmingly, these stronger models often exhibit higher susceptibility to Einstellung traps ($R_{bias}$ 51\%--61\%). For instance, \textbf{Gemini-2.5-Pro}, despite its high capability, shows a significantly higher bias rate (60.90\%) compared to \textbf{Claude-Sonnet-4.5} (42.98\%). This implies that in adversarial contexts, high capability can paradoxically increase vulnerability to bias. 
This result reveals a counter-intuitive conclusion: \textbf{current Scaling Laws enhance "statistical fitting" but fail to confer "differential diagnostic capability in dynamic contexts".}As corroborated by our failure mode analysis (Figure~\ref{fig:failure_modes}), highly capable models like GPT-5 exhibit a disproportionately high rate of \textit{Blindness}.This suggests that stronger models fit the prior distribution of training data so aggressively that they literally filter out low-probability counter-evidence during perception, making it structurally harder to escape the Einstellung Effect.

\paragraph{2. Existing Agents Amplify Cognitive Bias.}
Compared to the base model (Qwen3-32B), the multi-agent framework MDAgent does not yield the expected improvements and even exhibits degradation. We attribute this to two factors:
(1) \textbf{Noise Amplification:} The significant drop in $Acc_{base}$ (40.26\% $\to$ 29.70\%) suggests that without causal constraints, the diverse viewpoints introduced by multi-agent debate act as noise rather than signal.
(2) \textbf{Bias Amplification:} The stagnation in $Acc_{rob}$ and high $R_{bias}$ indicate that the "debate" mechanism, when faced with strong Einstellung traps, devolves into \textbf{Consensus Bias}, reinforcing the incorrect intuitive consensus rather than correcting it.

\paragraph{3. Effectiveness of Evidence-Based Architecture.}
In contrast, ECR-Agent (based on Qwen3-32B) achieves a qualitative leap in performance. It significantly boosts fundamental capability ($Acc_{base} \to$ 69.49\%) while doubling robustness ($Acc_{rob} \to$ 24.21\%) and reducing the bias rate ($R_{bias} \to$ 33.75\%). This demonstrates that introducing \textbf{Structural Causal Reasoning} and \textbf{Evidence Audit} mechanisms is key to breaking the Einstellung Effect. Unlike baselines that rely on internal parametric memory, ECR-Agent enforces an evidence-based reasoning process that prioritizes "evidence" over "probability," effectively circumventing the Einstellung Traps.

\subsection{Impact of Scale and Pathology}

\paragraph{Scaling Ineffectiveness.}
Figure~\ref{fig:scaling} visualizes the relationship between $R_{bias}$ and $Acc_{base}$. The results show no significant linear negative correlation, with data points widely scattered. Frontier models like GPT-5, despite possessing extreme fundamental capability (right side of X-axis), still exhibit very high bias rates (top of Y-axis). This indicates that reasoning robustness does not emerge naturally from scale. Without a structured verification mechanism, even advanced CoT reasoning remains susceptible to being trapped in the Einstellung Effect by strong statistical priors.

\paragraph{Pathology-Dependent Vulnerability.}
The clustering patterns in Figure~\ref{fig:scaling} and the heatmap in Figure~\ref{fig:heatmap} reveal the structural nature of the Einstellung Effect:
\begin{itemize}
    \item \textbf{Clustering:} Pathologies like \textit{Pneumonia} and \textit{Pericarditis} consistently appear in the High Bias Cluster across almost all models. This reveals strong \textbf{Spurious Correlations} in the training data.
    \item \textbf{Variance:} Conversely, pathologies like \textit{Influenza} show high variance, suggesting that when statistical priors are weaker, some models can successfully reason through distractors.
\end{itemize}
This pathology dependence confirms the systemic vulnerability of probabilistic models when facing "High-Confidence Prior vs. Low-Confidence Evidence" conflicts. ECR-Agent succeeds by transforming the "probability prediction problem" into an "evidence verification problem" via Causal Intervention, structurally blocking the propagation of spurious correlations.

\section{Case Study}
\label{app:case_study}

To demonstrate the efficacy of \textsc{MedEinst} in benchmarking the Einstellung Effect and the robustness of \textsc{ECR-Agent}, we present a detailed analysis of \textbf{Case 100473}. This case represents a high-stakes emergency scenario where the baseline model succumbed to a "Pattern Matching" trap, while our agent successfully corrected the diagnosis through causal graph reasoning.

\subsection{Case Overview}
\begin{itemize}
    \item \textbf{Ground Truth:} Pulmonary Embolism (PE).
    \item \textbf{Trap Type:} \textit{Distractor Injection} (Family History of Pneumothorax) + \textit{Evidence Substitution} (History of DVT).
    \item \textbf{Baseline Intuition:} Spontaneous Pneumothorax.
    \item \textbf{ECR-Agent Verdict:} Overturn $\to$ Pulmonary Embolism.
\end{itemize}

\begin{table*}[h!]
    \centering
    \small
    \begin{tabularx}{\textwidth}{@{}l X X@{}}
        \toprule
        \textbf{Feature} & \textbf{Control Case ($x^c$)} & \textbf{Trap Case ($x^t$)} \\
        \midrule
        \textbf{Demographics} & Male, 22 years old & Male, 22 years old \\
        \textbf{Chief Complaint} & Sudden "knife-like" chest pain, Dyspnea & Sudden "knife-like" chest pain, Dyspnea \\
        \textbf{Key Evidence} & History of \textbf{Spontaneous Pneumothorax} & History of \textbf{Deep Vein Thrombosis (DVT)} \\
        \textbf{Distractors} & Family history of Pneumothorax & Family history of Pneumothorax \\
        \textbf{Associated Sx} & Tachycardia, Hypoxia & Tachycardia, Hypoxia \\
        \midrule
        \textbf{Model Diagnosis} & Spontaneous Pneumothorax (\checkmark) & Spontaneous Pneumothorax (\textbf{Error}) \\
        \bottomrule
    \end{tabularx}
    \caption{Comparison of the Control and Trap narratives. The \textbf{Trap Case} replaces the patient's personal history with DVT (a risk factor for PE) but retains the family history of Pneumothorax, triggering the Einstellung Effect in baseline models.}
    \label{tab:case_narrative}
\end{table*}

\subsection{Narrative Comparison}
Table \ref{tab:case_narrative} illustrates the minimal yet critical differences between the Control and Trap cases. The Trap case introduces a strong "Red Herring" (Family History) while subtly embedding the key discriminative evidence (DVT History).

\subsection{Reasoning Trace Analysis}
The baseline model (intuition) anchored on the "Young Male + Sudden Chest Pain + Family History" pattern, incorrectly diagnosing Spontaneous Pneumothorax. Below is the \textbf{reconstructed audit log} from the \textsc{ECR-Agent}'s \textit{Evidence Audit}, demonstrating how it utilized the Causal Graph to overturn this error.

\begin{quote}
\textbf{Evidence Audit Log:} \\
\textit{Initial Hypothesis:} Spontaneous Pneumothorax (Probability: High). \\
\textbf{Tier 1: Fatal Conflict Check} \\
- \textit{Check:} Pneumothorax typically requires specific imaging confirmation (e.g., "Spontaneous pneumothorax on imaging"). \\
- \textit{Result:} \textcolor{red}{Missing/Shadow Node}. The narrative does not mention imaging evidence. \\
\textbf{Tier 2: Pivot Competition} \\
- \textit{Pivot 1 (DVT History):} Strongly supports \textbf{Pulmonary Embolism} (Risk Factor). \textcolor{blue}{Matched} in Patient Narrative. \\
- \textit{Pivot 2 (Family History):} Supports Pneumothorax, but is a weak "General" feature compared to the specific risk factor of DVT. \\
- \textit{Pivot 3 (Hypoxia + Tachycardia):} Supports both, but biologically more severe in PE. \\
\textbf{Decision:} The presence of "History of DVT" is a distinct Pivot Node that rules out Pneumothorax (as a primary cause) and strongly supports PE. The initial intuition was biased by the family history. \\
\textbf{Final Verdict:} \textbf{OVERTURN} $\to$ Pulmonary Embolism.
\end{quote}

\subsection{Interpretability: Evidence Balance Sheet}
The core of \textsc{ECR-Agent}'s interpretability lies in its explicit \textbf{Causal Graph}. Table \ref{tab:evidence_balance} details the "Evidence Balance Sheet" for Case 100473.

The agent constructs a graph connecting the Patient Observations ($P_{obs}$) to the Knowledge Nodes ($K_{nodes}$) of competing diagnoses. The decision is driven by \textbf{Pivot Nodes}—features that logically distinguish between the two conditions.

\begin{table*}[h!]
    \centering
    \small
    \begin{tabularx}{\textwidth}{l l l X}
        \toprule
        \textbf{Diagnosis Candidate} & \textbf{Node Type} & \textbf{Relation to Patient} & \textbf{Clinical Feature Content} \\
        \midrule
        \multicolumn{4}{l}{\textit{\textbf{Diagnosis A: Pulmonary Embolism (Correct)}}} \\
        & \textbf{Pivot} & \textcolor{green}{\textbf{Match}} & \textbf{History of Deep Vein Thrombosis (DVT)} \\
        & \textbf{Pivot} & \textcolor{green}{\textbf{Match}} & Sudden onset dyspnea with tachycardia \& hypoxia \\
        & \textbf{Pivot} & \textcolor{green}{\textbf{Match}} & Sharp pleuritic chest pain exacerbated by inspiration \\
        & General & Match & Dyspnea with sudden onset \\
        \midrule
        \multicolumn{4}{l}{\textit{\textbf{Diagnosis B: Spontaneous Pneumothorax (Intuition/Error)}}} \\
        & \textbf{Pivot} & \textcolor{red}{\textbf{Missing}} & Spontaneous pneumothorax on imaging \\
        & General & Match & History of spontaneous pneumothorax \& family history \\
        & Pivot & Match & Acute onset pleuritic chest pain \\
        & \textbf{Pivot} & \textcolor{orange}{\textbf{Rule Out}} & \textit{Presence of prior Deep Vein Thrombosis (DVT)} \\
        \bottomrule
    \end{tabularx}
    \caption{\textbf{Evidence Balance Sheet.} The table shows why the agent favored PE over Pneumothorax. While Pneumothorax has matching symptoms (chest pain), it lacks its critical Pivot evidence (Imaging) and is actively ruled out by the presence of DVT, which is a Pivot Match for PE.}
    \label{tab:evidence_balance}
\end{table*}

\section{Extended Discussion: Theoretical Grounding and Comparative Analysis}
\label{app:extended_discussion}

While the main text outlines the broad landscape of medical LLMs, this appendix provides a deeper theoretical analysis of why existing paradigms—specifically Multi-Agent Collaboration and Retrieval-Augmented Generation (RAG)—insufficiently address the Einstellung Effect, and how our \textsc{ECR-Agent} fundamentally differs by aligning with Causal Inference theories.

\subsection{Verification vs. Consensus: The Limits of Multi-Agent Debate}
Recent agentic frameworks like MDAgents \citep{kim2024mdagents} and MedAgents \citep{medagents} rely on "collaboration" or "debate" strategies, assuming that diverse personas will cancel out individual errors. However, this assumption holds only when errors are independent and randomly distributed. 

In the context of the \textbf{Einstellung Effect}, errors are not random but \textit{systematic}. As shown in our experiments (Table 1), strong statistical priors act as a "common distractor" that misleads the majority of models/agents similarly.
\begin{itemize}
    \item \textbf{Consensus Bias:} When the "intuitive but wrong" diagnosis is statistically dominant, multi-agent debate often devolves into \textbf{Consensus Bias} \citep{biasmedqa,schmidgall2024consensus}. Agents tend to converge on the most likely probabilistic token rather than the ground truth evidence.
    \item \textbf{Our Solution (Veto by Evidence):} Unlike debate frameworks that optimize for \textit{agreement}, \textsc{ECR-Agent} optimizes for \textit{falsification}. By introducing \textbf{Pivot Nodes} (Section 4.2.2), our agent grants a single piece of discriminative evidence the power to "veto" the majority consensus, mirroring the clinical principle that "one proven contradiction outweighs a thousand probabilities".
\end{itemize}

\subsection{Dynamic Inference vs. Static Knowledge: The Limits of RAG}
Retrieval-Augmented Generation (RAG) systems, such as MedGraphRAG \citep{medrag} and PrimeKG \citep{primekg}, attempt to mitigate hallucinations by retrieving external knowledge. While effective for factual queries, standard RAG faces structural limitations in \textbf{Counterfactual Differential Diagnosis}:
\begin{itemize}
    \item \textbf{Static vs. Dynamic:} RAG retrieves \textit{static} associations (e.g., "Pulmonary Embolism causes Chest Pain") but lacks the mechanism to construct a \textit{patient-specific} causal graph. It cannot dynamically evaluate "What if this specific symptom was absent?" or "Why is this overlapping symptom non-discriminative in this specific context?".
    \item \textbf{Associative vs. Causal:} RAG fundamentally enhances \textit{Associative Reasoning} (Pearl's Layer 1) by adding more context to the prompt. It does not perform \textit{Intervention} (Layer 2). 
    \item \textbf{Our Solution:} \textsc{ECR-Agent} does not just retrieve knowledge; it structures it into a \textbf{Dynamic Causal Graph}. By explicitly modeling \textit{Match}, \textit{Conflict}, and \textit{Shadow} relations, we transform static knowledge into active reasoning tools that can perform logical interventions on the patient's narrative.
\end{itemize}

\subsection{Theoretical Grounding: Mapping Diagnosis to the Causal Hierarchy}
Our framework is theoretically grounded in the integration of Evidence-Based Medicine (EBM) \citep{EBM} with Pearl’s Causal Hierarchy \citep{pearl2018book}. We provide a formal mapping of these cognitive processes:

\begin{enumerate}
    \item \textbf{Layer 1: Association.} 
    \textit{Clinical Equivalent:} Pattern Recognition / Intuition.
    \textit{Implementation:} Our \textbf{Dual-Pathway Perception} module generates initial hypotheses based on $P(Diagnosis | Symptoms)$. This is where the Einstellung Effect (statistical bias) originates.
    
    \item \textbf{Layer 2: Intervention.} 
    \textit{Clinical Equivalent:} Differential Diagnosis / Testing.
    \textit{Implementation:} Our \textbf{Forward Causal Reasoning} simulates the act of "intervening" to find truth. We define \textbf{Pivot Nodes} as the minimal intervention set $do(X)$ required to distinguish between competing hypotheses $d_i$ and $d_j$. This aligns with \citet{causalml}, who proved that optimal diagnosis requires maximizing the Information Gain of interventions.
    
    \item \textbf{Layer 3: Counterfactuals.} 
    \textit{Clinical Equivalent:} Diagnostic Verification / Audit.
    \textit{Implementation:} Our \textbf{Backward Causal Reasoning} and \textbf{Evidence Audit} perform the counterfactual check: "Given diagnosis $d$, what symptom $s$ \textit{would have been} observed?". The detection of \textbf{Shadow Nodes} (missing expected evidence) formally represents the violation of counterfactual expectations ($P(s_{missing} | do(d)) \approx 0$), allowing the model to reject high-probability but causally inconsistent traps.
\end{enumerate}

This rigorous mapping demonstrates that \textsc{ECR-Agent} is not merely an engineering improvement but a step towards \textbf{Causal AI} in medicine, moving beyond the \textit{Curve Fitting} limitations of standard LLMs \citep{causalml}.

\section{Data Samples}
\label{app:data_samples}

To demonstrate the realistic clinical presentation of \textsc{MedEinst}, Figure \ref{fig:narrative_comparison} displays the raw input narratives for Case 100473 as they appear to the model.

We adopt the structured format from the DDXPlus dataset, which organizes clinical observations into \texttt{Symptoms} and \texttt{Antecedents} with hierarchical indentation. The figure highlights the \textbf{counterfactual intervention}: while the lengthy symptom description and the "Family history" distractor remain identical, the specific patient history in the \texttt{Antecedents} section is surgically altered from "Spontaneous pneumothorax" (Control) to "Deep vein thrombosis" (Trap).

\begin{figure*}[p] 
    \centering
    \renewcommand{\ttdefault}{pcr} 
    \footnotesize
    
    \begin{minipage}[t]{0.46\textwidth}
        \textbf{Control Case ($x^c$): Spontaneous Pneumothorax}
        \hrule
        \vspace{1mm}
\begin{verbatim}
Sex: M, Age: 22
Geographical region: North America

Symptoms:
---------
- I feel pain.
  - The pain is:
    * heartbreaking
    * a knife stroke
  - The pain locations are:
    * side of the chest(R)
    * breast(R)
    * breast(L)
- On a scale of 0-10, 
  the pain intensity is 6
- The pain radiates to 
  these locations:
  * nowhere
- On a scale of 0-10, 
  the location precision is 2
- On a scale of 0-10, 
  the speed of onset is 9
- I am experiencing shortness 
  of breath or 
  difficulty breathing in a significant way.
- I have pain that is increased 
  when I breathe 
  in deeply.
- I have tachycardia.
- I have hypoxia.

Antecedents:
------------
- I have had a spontaneous pneumothorax.
- I smoke cigarettes.
- One or more of my family members have had 
  a pneumothorax.
- I have not traveled out of the country in 
  the last 4 weeks.
\end{verbatim}
    \end{minipage}
    \hfill
    \begin{minipage}[t]{0.46\textwidth}
        \textbf{Trap Case ($x^t$): Pulmonary Embolism}
        \hrule
        \vspace{1mm}
\begin{verbatim}
Sex: M, Age: 22
Geographical region: North America

Symptoms:
---------
- I feel pain.
  - The pain is:
    * heartbreaking
    * a knife stroke
  - The pain locations are:
    * side of the chest(R)
    * breast(R)
    * breast(L)
- On a scale of 0-10, 
  the pain intensity is 6
- The pain radiates to 
  these locations:
  * nowhere
- On a scale of 0-10, 
  the location precision is 2
- On a scale of 0-10, 
  the speed of onset is 9
- I am experiencing shortness 
  of breath or 
  difficulty breathing in a significant way.
- I have pain that is increased 
  when I breathe 
  in deeply.
- I have tachycardia.
- I have hypoxia.

Antecedents:
------------
- I have had a deep vein thrombosis (DVT). <!!>
- I smoke cigarettes.
- One or more of my family members have had 
  a pneumothorax.
- I have not traveled out of the country in 
  the last 4 weeks.
\end{verbatim}
    \end{minipage}
    
    \caption{Side-by-side comparison of the raw clinical narratives for Case 100473. The text is presented in the original DDXPlus format used as input for the LLMs. The \textbf{Trap Case} (Right) contains a minimal edit in the \texttt{Antecedents} section (marked with \texttt{<!!>}), replacing the history of pneumothorax with DVT, while retaining the misleading family history.}
    \label{fig:narrative_comparison}
\end{figure*}

\section{Prompts Details}
\label{sec:appendix_prompts}

To ensure the reproducibility of our work, we provide the full system prompts used in both the \textsc{MedEinst} benchmark construction pipeline and the \textsc{ECR-Agent} reasoning framework.

\subsection{MedEinst Benchmark Construction}
Tables \ref{tab:prompt_extract}, \ref{tab:prompt_trap}, \ref{tab:prompt_rewrite}, and \ref{tab:prompt_judge} detail the prompts for the four-stage adversarial data construction pipeline.

\begin{table*}[h!]
    \centering
    \small
    \begin{tabularx}{\textwidth}{X}
        \toprule
        \textbf{Discriminative Feature Extraction Prompt (Step 1)} \\
        \midrule
        \texttt{System Prompt:} You are a senior medical expert. Your task is to perform a differential diagnosis based on the provided reference knowledge. \\
        \textbf{CRITICAL INSTRUCTION:} You MUST use the 'Reference Knowledge' below as your ONLY source of truth. Do not use your own internal knowledge. \\
        \textbf{Reference Knowledge:} \\
        - Typical Symptoms of '{{distractor\_disease}}': {{distractor\_symptoms}} \\
        - Typical Symptoms of '{{truth\_disease}}': {{truth\_symptoms}} \\
        \textbf{Patient Narrative:} --- {{control\_narrative}} --- \\
        \textbf{Your Task:} \\
        1. Compare the patient's narrative with the two lists of typical symptoms. \\
        2. Identify the SINGLE MOST CRITICAL and ATOMIC phrase within the narrative that is a typical symptom of '{{distractor\_disease}}' but NOT a typical symptom of '{{truth\_disease}}'. \\
        \textbf{Response Format:} If you successfully identify such a phrase, respond in JSON format with one key: \{"narrative\_A": "the exact phrase from the narrative"\}. \\
        \bottomrule
    \end{tabularx}
    \caption{Prompt used to extract the key discriminative evidence ($k_{gt}$) that supports the control diagnosis.}
    \label{tab:prompt_extract}
\end{table*}

\begin{table*}[h!]
    \centering
    \small
    \begin{tabularx}{\textwidth}{X}
        \toprule
        \textbf{Trap Information Generation Prompt (Step 2)} \\
        \midrule
        \texttt{System Prompt:} You are a medical writer. Your task is to generate an 'isomorphic' clinical finding, grounded in the provided reference knowledge. \\
        \textbf{CRITICAL INSTRUCTION:} You MUST choose an evidence from the 'Reference Knowledge for {{truth\_disease}}' list below. \\
        \textbf{Reference Knowledge for {{truth\_disease}}:} ... \\
        \textbf{The original phrase (pointing to {{distractor\_disease}}):} "{{narrative\_A}}" \\
        \textbf{Your Task:} \\
        1. Review the list of typical evidences for '{{truth\_disease}}'. \\
        2. Select one evidence from those lists that is most 'isomorphic' to the original phrase in terms of clinical gravity and role. \\
        3. Rephrase the selected evidence into a concise, ATOMIC phrase a patient would say. \\
        Respond in JSON format with one key: "narrative\_B". \\
        \bottomrule
    \end{tabularx}
    \caption{Prompt used to generate the misleading trap feature ($k_{trap}$) based on the bias disease knowledge.}
    \label{tab:prompt_trap}
\end{table*}

\begin{table*}[h!]
    \centering
    \small
    \begin{tabularx}{\textwidth}{X}
        \toprule
        \textbf{Differential Features Rewrite Prompt (Step 3)} \\
        \midrule
        \texttt{System Prompt:} You are an expert medical writer performing a "Cognitive Surgery". Your task is to seamlessly rewrite a clinical narrative to change its diagnostic direction. \\
        \textbf{Original Patient Narrative:} --- {{control\_narrative}} --- \\
        \textbf{Your instructions:} \\
        1. Locate the phrase "{{narrative\_A}}" in the text. \\
        2. Rewrite the single sentence containing this phrase to instead convey the new core information: "{{narrative\_B}}". \\
        3. \textbf{Crucially, you MUST ensure the new sentence is grammatically perfect and logically fits the surrounding context.} The style and tone must match the original exactly. \\
        4. \textbf{Do not change any other part of the narrative.} \\
        Respond with only the complete, rewritten patient narrative text. \\
        \bottomrule
    \end{tabularx}
    \caption{Prompt used to inject the trap feature into the patient narrative ($x^c \rightarrow x^t$).}
    \label{tab:prompt_rewrite}
\end{table*}

\begin{table*}[h!]
    \centering
    \small
    \begin{tabularx}{\textwidth}{X}
        \toprule
        \textbf{Inter-Model Verification Prompt (Step 4)} \\
        \midrule
        \texttt{System Prompt:} You are an expert clinical diagnostician and medical narrative analyst, acting as an impartial judge. Your task is to assess the quality of a synthetically modified "Trap Case". \\
        \textbf{Your evaluation must be based on three precise criteria:} \\
        1. \textbf{Diagnostic Correctness (Boolean):} When reading the trap case narrative independently, does it provide sufficient evidence to make its stated ground truth ('{{trap\_gt}}') a plausible and likely diagnosis? \\
        2. \textbf{Medical Plausibility (1-10 Scale):} How medically believable is the complete trap case narrative? \\
        3. \textbf{Narrative Fluency (1-10 Scale):} How well was the new information integrated? \\
        ... \\
        \bottomrule
    \end{tabularx}
    \caption{LLM-as-a-Judge prompt for verifying the quality and validity of generated trap cases.}
    \label{tab:prompt_judge}
\end{table*}

\subsection{ECR-Agent Reasoning Framework}
Tables \ref{tab:prompt_phase1}, \ref{tab:prompt_phase2}, and \ref{tab:prompt_phase3} detail the prompts for the three-phase causal reasoning engine.

\begin{table*}[h!]
    \centering
    \small
    \begin{tabularx}{\textwidth}{X}
        \toprule
        \textbf{Analytic Problem Representation Prompt (Dual-Pathway Perception)} \\
        \midrule
        \texttt{System Prompt:} You are a Senior Clinical Diagnostician and Expert Medical Scribe. Your task is to perform "Problem Representation" on a raw patient case. \\
        \textbf{OBJECTIVE:} Transform the patient's raw narrative into a structured list of \textbf{P-Nodes (Patient Features)} using precise \textbf{Medical Semantic Qualifiers}. \\
        \textbf{THE PROCESS:} \\
        1. Translate Time ... 2. Translate Symptoms ... 3. Filter ... 4. Synthesize ... \\
        \textbf{OUTPUT SCHEMA (JSON):} \\
        Return a single JSON object with two keys: `problem\_representation\_one\_liner` and `p\_nodes` (containing id, content, original\_text, status). \\
        \textbf{RULES:} Only mark status: "Absent" if the text explicitly says "no", "denies", or "without". \\
        \bottomrule
    \end{tabularx}
    \caption{Prompt for extracting structured patient observations ($P_{obs}$) from raw text.}
    \label{tab:prompt_phase1}
\end{table*}

\begin{table*}[h!]
    \centering
    \small
    \begin{tabularx}{\textwidth}{X}
        \toprule
        \textbf{Pivot Node Discovery Prompt (Causal Graph Reasoning)} \\
        \midrule
        \texttt{System Prompt:} You are an Expert Diagnostician performing a comprehensive Differential Diagnosis. \\
        \textbf{Step 1: Disease-by-Disease Analysis} ... \\
        \textbf{Step 2: Cross-Disease Comparison (Matrix Analysis)} \\
        Create a mental discrimination matrix: Which features are UNIQUE to one disease? Which features RULE OUT certain diseases? \\
        \textbf{OUTPUT JSON SCHEMA:} \\
        You MUST output a JSON object with: "k\_nodes": [ \{ "content": "...", "type": "Pivot", "importance": "Pathognomonic", "supported\_candidates": [...], "ruled\_out\_candidates": [...] \} ] \\
        \textbf{FIELD DEFINITIONS:} \\
        - \textbf{Pivot}: Discriminating feature that helps distinguish between 2+ diseases. \\
        \bottomrule
    \end{tabularx}
    \caption{Prompt for identifying Pivot Nodes to differentiate between competing hypotheses.}
    \label{tab:prompt_phase2}
\end{table*}

\begin{table*}[h!]
    \centering
    \small
    \begin{tabularx}{\textwidth}{X}
        \toprule
        \textbf{Evidence Audit \& Final Decision Prompt (Evidence Audit)} \\
        \midrule
        \texttt{System Prompt:} You are the \textbf{Chief Medical Auditor} and Final Decision Maker. Your goal is to audit the reasoning of a "System 1" (Initial Intuition) agent using a "System 2" (Causal Graph) evidence map. \\
        \textbf{THE LOGIC HIERARCHY (Follow Strictly)} \\
        \textbf{Tier 1: The Safety Sentinel (Fatal Conflicts)} \\
        Rule: If a Candidate requires a symptom that is `Essential`, but the Patient explicitly has `Status: Absent`, then this Candidate is DISQUALIFIED. \\
        \textbf{Tier 2: The Pivot Competition (Differential Diagnosis)} \\
        Rule: A Candidate supported by a \textbf{matched Pivot Feature} is superior to a Candidate supported only by General features. \\
        \textbf{Tier 3: The Shadow \& Coverage Audit (Tie-Breaker)} \\
        Select the candidate with the highest explanatory coverage and fewest unexplained conflicts. \\
        \bottomrule
    \end{tabularx}
    \caption{Prompt for the final evidence audit, applying the Tiered Logic Hierarchy to select the diagnosis.}
    \label{tab:prompt_phase3}
\end{table*}

\end{document}